%% file: metricPaper_Bylinskii_PAMI_withappendix.tex
\documentclass[10pt,journal,cspaper,compsoc]{IEEEtran}
\usepackage{graphicx}
\usepackage{url}
\usepackage{caption}
\usepackage{colortbl}
\usepackage{amssymb}
\usepackage{amsmath}

\usepackage{booktabs,xcolor,siunitx}
\definecolor{lightgray}{gray}{0.9}
\definecolor{lightblue}{gray}{0.6}

\ifCLASSOPTIONcompsoc
\else
   \usepackage{cite}
\fi
\ifCLASSINFOpdf
\else
\fi

\begin{document}
%
\title{What do different evaluation metrics\\ tell us about saliency models?}

%
%
%
%

\author{{\normalsize 
Zoya~Bylinskii*,
 Tilke~Judd*,
        Aude~Oliva,
        Antonio~Torralba,
        and~Fr\'edo~Durand}
\IEEEcompsocitemizethanks{\IEEEcompsocthanksitem Zoya Bylinskii, Aude Oliva, Antonio Torralba, and Fr\'edo Durand are with the Computer Science and Artificial Intelligence Laboratory, Massachusetts Institute of Technology, Cambridge,
MA, 02139.  E-mail: \{zoya, oliva, torralba, fredo\}@csail.mit.edu. 
\newline Tilke Judd (tilke.judd@gmail.com) is at Google, Zurich.
\newline * indicates equal contribution.
}
} 

\IEEEcompsoctitleabstractindextext{%
\begin{abstract}
How best to evaluate a saliency model's ability to predict where humans look in images is an open research question. The choice of evaluation metric depends on how saliency is defined and how the ground truth is represented. Metrics differ in how they rank saliency models, and this results from how false positives and false negatives are treated, whether viewing biases are accounted for, whether spatial deviations are factored in, and how the saliency maps are pre-processed. In this paper, we provide an analysis of 8 different evaluation metrics and their properties. With the help of systematic experiments and visualizations of metric computations, we add interpretability to saliency scores and more transparency to the evaluation of saliency models. Building off the differences in metric properties and behaviors, we make recommendations for metric selections under specific assumptions and for specific applications.
\end{abstract}

}


\maketitle

\IEEEdisplaynotcompsoctitleabstractindextext

%
\IEEEpeerreviewmaketitle

\section{Introduction}

Automatically predicting regions of high saliency in an image is useful for applications including content-aware image re-targeting, image compression and progressive transmission, object and motion detection, image retrieval and matching. Where human observers look in images is often used as a ground truth estimate of image saliency, and computational models producing a saliency value at each pixel of an image are referred to as saliency models\footnote{Although the term saliency was traditionally used to refer to bottom-up conspicuity, many modern saliency models include scene layout, object locations, and other contextual information.}.

\begin{figure*}
\centering
\includegraphics[width=0.9\linewidth]{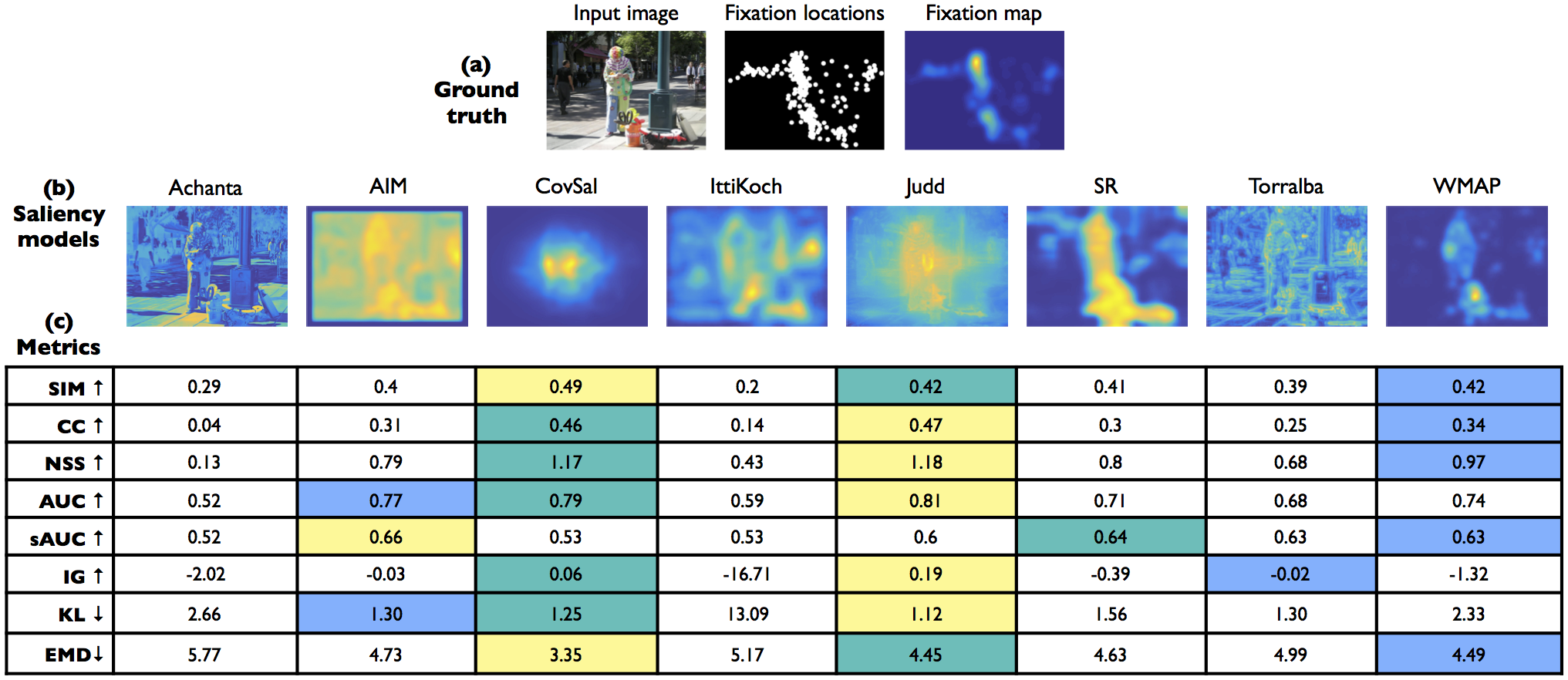}
\caption{\small{Evaluation metrics score saliency models differently. Saliency maps are evaluated on how well they approximate human ground truth eye movements, represented either as discrete fixation locations or a continuous fixation map (a). For a given image, saliency maps corresponding to 8~saliency models (b) are scored under 8 different evaluation metrics (6 similarity and 2 dissimilarity metrics), highlighting the top 3 best scoring maps under each metric for this particular image~(c). }}
\label{fig:main}
\vspace{-1em}
\end{figure*}

Dozens of computational saliency models are available to choose from~\cite{borji2013quantitative,BorjiICCV2013,BylinskiiOpinion,mit-saliency-benchmark,Judd_2012}, but objectively determining which model offers the ``best" approximation to human eye fixations remains a challenge.
For example, for the input image in Fig.~\ref{fig:main}a, we include the output of 8 different saliency models (Fig.~\ref{fig:main}b). When compared to human ground truth the saliency models receive different scores according to different evaluation metrics (Fig.~\ref{fig:main}c).  
The inconsistency in how different metrics rank saliency models can often leave performance up to interpretability.

In this paper, we quantify metric behaviors. Through a series of systematic experiments and novel visualizations (Fig.~\ref{fig:vis}), we aim to understand how changes in the input saliency maps impact metric scores, and as a result why models are scored differently.  
Some metrics take a probabilistic approach to distribution comparison, yet others treat distributions as histograms or random variables (Sec.~\ref{sec:metrics}). Some metrics are especially sensitive to false negatives in the input prediction, others to false positives, center bias, or spatial deviations (Sec.~\ref{sec:discussion}). 
Differences in how saliency and ground truth are represented and which attributes of saliency models should be rewarded/penalized leads to different choices of metrics for reporting performance \cite{BorjiICCV2013,mit-saliency-benchmark,kummerer2014close,kummerer2015information,LeMeur2013,RicheICCV2013,wilming2011measures}. 
We consider metric behaviors in isolation from any post-processing or regularization on the part of the models. 

Building on the results of our analyses, we offer guidelines for designing saliency benchmarks (Sec.~\ref{sec:designSalBenchmark}). 
For instance, for evaluating probabilistic saliency models we suggest the KL-divergence and Information Gain (IG) metrics. For benchmarks like the MIT Saliency Benchmark which do not expect saliency models to be probabilistic, but do expect models to capture viewing behavior including systematic biases, we recommend  either Normalized Scanpath Saliency (NSS) or Pearson's Correlation Coefficient (CC).

Our contributions include:
\begin{itemize}
\item An analysis of 8 metrics commonly used in saliency evaluation. We discuss how these metrics are affected by different properties of the input, and the consequences for saliency evaluation. 
\item Visualizations for all the metrics to add interpretability to metric scores and transparency to the evaluation of saliency models.
\item An accompanying manuscript to the MIT Saliency Benchmark to help interpret results.
\item Guidelines for designing new saliency benchmarks, including defining expected inputs and modeling assumptions, specifying a target task, and choosing how to handle dataset bias.
\item Advice for choosing saliency evaluation metrics based on design choices and target applications.
\end{itemize}

\section{Related work}
\label{sec:related}

\subsection{Evaluation metrics for computer vision}

Similarity metrics operating on image features have been a subject of investigation and application to different computer vision domains \cite{liu2007survey,smeulders2000content,vasconcelos2004efficient,zitova2003image}. Images are often represented as histograms or distributions of features, including low-level features like edges (texture), shape and color, and higher-level features like objects, object parts, and bags of low-level features. Similarity metrics applied to these feature representations have been used for classification, image retrieval, and image matching tasks \cite{RubnerTomasi,sinha2013study,smeulders2000content}. Properties of these metrics across different computer vision tasks also apply to the task of saliency modeling, and we provide a discussion of some applications in Sec.~\ref{sec:applications}. 
The discussion and analysis of the metrics in this paper can correspondingly be generalized to other computer vision applications.

\subsection{Evaluation metrics for saliency}
\label{sec:relatedwork_metrics}

A number of papers in recent years have compared models across different metrics and datasets.
Wilming et al.~\cite{wilming2011measures} discussed the choice of metrics for saliency model evaluation, deriving a set of qualitative and high-level desirable properties for metrics: ``few parameters", ``intuitive scale", ``low data demand", and ``robustness". Metrics were discussed from a theoretical standpoint without empirical experiments or quantification of metric behavior. 

Le Meur and Baccino~\cite{LeMeur2013} reviewed many methods of comparing scanpaths and saliency maps. For evaluation, however, only 2 metrics were used to compare 4 saliency models.
Sharma and Alsam \cite{tseng2009quantifying} reported the performance of 11 models with 3 versions of the AUC metric on MIT1003 \cite{Judd_2009}.
Zhao and Koch~\cite{Zhao10032011} performed an analysis of saliency on 4 datasets using 3 metrics.
Riche et al.~\cite{RicheICCV2013} provided an evaluation 12 saliency models with 12 similarity metrics on Jian Li's dataset \cite{Li2012}. 
They compared how metrics rank saliency models and reported which metrics cluster together, but did not provide explanations.

\begin{table}[h!]
\centering
\begin{tabular}{|p{3cm}|p{1cm}|p{2.5cm}|}
\hline
Metric & Denoted here & Evaluation papers appearing in \\ \hline
Area under ROC Curve & AUC & \cite{BorjiICCV2013,emami2013selection,engelke2013comparative,li2015data,LeMeur2013,RicheICCV2013,wilming2011measures,Zhao10032011} \\ \hline
Shuffled AUC & sAUC & \cite{borji2013quantitative,BorjiICCV2013,li2015data,RicheICCV2013} \\ \hline
Normalized Scanpath Saliency & NSS & \cite{borji2013quantitative,BorjiICCV2013,emami2013selection,li2015data,LeMeur2013,RicheICCV2013,wilming2011measures,Zhao10032011} \\ \hline
Pearson's Correlation Coefficient & CC & \cite{borji2013quantitative,BorjiICCV2013,emami2013selection,engelke2013comparative,li2015data,RicheICCV2013,wilming2011measures} \\ \hline
Earth Mover's Distance & EMD & \cite{li2015data,RicheICCV2013,Zhao10032011} \\ \hline
Similarity or histogram intersection &SIM & \cite{li2015data,RicheICCV2013} \\ \hline
Kullback-Leibler divergence & KL & \cite{emami2013selection,li2015data,RicheICCV2013,wilming2011measures} \\ \hline
Information Gain & IG & \cite{kummerer2014close,kummerer2015information} \\ \hline
\end{tabular}
\caption{{\small The most common metrics for saliency model evaluation are analyzed in this paper. We include a list of the surveys that have used these metrics.}}
\label{tab:metrics_used}
\vspace{-1em}
\end{table}

Borji, Sihite et al.~\cite{borji2013quantitative} compared 35 models on a number of image and video datasets 
using 3 metrics. Borji, Tavakoli et al.~\cite{BorjiICCV2013} compared 32 saliency models with 3 metrics for fixation prediction and additional metrics for scanpath prediction on 4 datasets. The effects of center bias and map smoothing on model evaluation were discussed. A synthetic experiment was run with a single set of random fixations 
while blur sigma, center bias, and border size were varied to determine how the 3 different metrics are affected by these transformations. Our analysis extends to 8 metrics tested on different variants of synthetic data to explore the space of metric behaviors.

Li et al.~\cite{li2015data} used crowdsourced perceptual experiments to discover which metrics most closely correspond to visual comparison of spatial distributions. Participants were asked to select out of pairs of saliency maps the map perceived to be closest to the ground truth map. Human annotations were used to order saliency models, and this ranking was compared to rankings by 9 different metrics. 
However, human perception can naturally favor some saliency map properties over others (Sec.~\ref{sec:qualitative_eval}). Visual comparisons are affected by the range and scale of saliency values, and are driven by the most salient locations, while small values are not as perceptible and don't enter into the visual calculations. This is in contrast to metrics that are particularly sensitive to zero values and regularization, which might nevertheless be more appropriate for certain applications, for instance when evaluating probabilistic saliency models (Sec.~\ref{sec:selectingMetrics}).

Emami and Hoberock~\cite{emami2013selection} compared 9 evaluation metrics (3 novel, 6 previously-published) in terms of human consistency. They defined the best evaluation metric as the one which best discriminates between a human saliency map and a random saliency map, as compared to the ground truth map. Human fixations were split into 2 sets, to generate human saliency maps and ground truth maps for each image. 
This procedure was the only criterion by which metrics were evaluated, and the chosen evaluation metric was used to compare 10 saliency models.

In this paper, we analyze metrics commonly used in other evaluation efforts (Table~\ref{tab:metrics_used}) and reported on the MIT Saliency Benchmark \cite{mit-saliency-benchmark}. 
We include Information Gain (IG), recently introduced by K{\"u}mmerer et al.~\cite{kummerer2014close,kummerer2015information}. 
To visualize metric computations and highlight differences in metric behaviors, we used standard saliency models for which code is available online. These models, depicted in Fig.~\ref{fig:main}b, include Achanta \cite{Achanta2009CVPR}, AIM \cite{Bruce13032009}, CovSal \cite{ErdemJoV2013}, IttiKoch \cite{Koch1985,Walther2006}, Judd \cite{Judd_2009}, SR \cite{Seo2012}, Torralba \cite{Torralba:2006}, and WMAP \cite{LopezGarcia2011}. Models were used for visualization purposes only, as the primary focus of this paper is comparing the metrics, not the models.

\begin{figure*}
\centering
\includegraphics[width=0.9\linewidth]{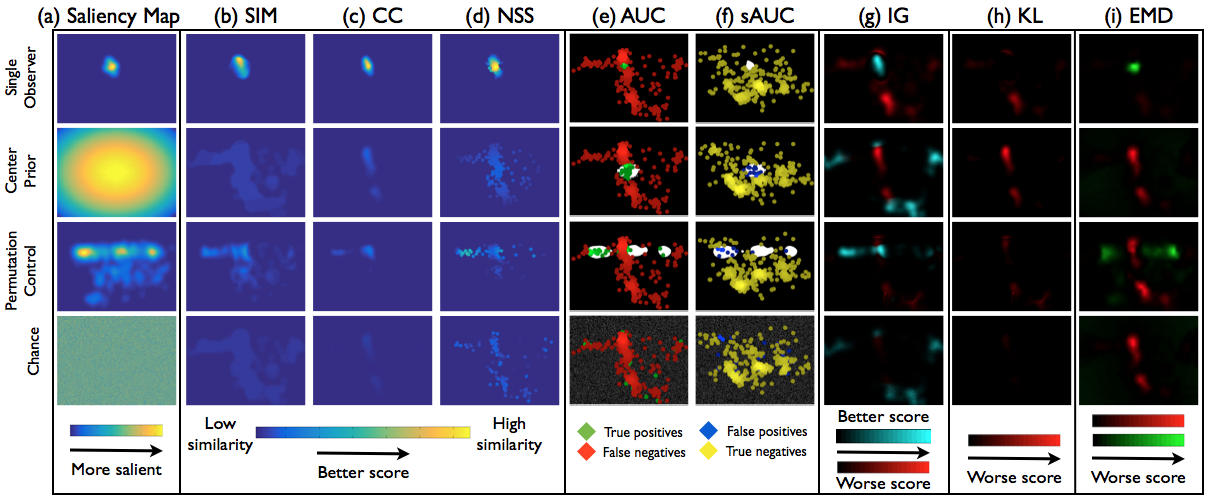}
\caption{\small{A series of experiments and corresponding visualizations can help us understand what behaviors of saliency models different evaluation metrics capture. Given a natural image and ground truth human fixations on the image as in Fig.~\ref{fig:main}a, we evaluate saliency models, including the 4 baselines in column (a), at their ability to approximate ground truth. Visualizations of 8 common metrics (b-i) help elucidate the computations performed when scoring saliency models. }}
\label{fig:vis}
\vspace{-1em}
\end{figure*}

Rather than providing tables of performance values and literature reviews of metrics, this paper offers intuition about how metrics perform under various conditions and where they differ, using experiments with synthetic and natural data, and visualizations of metric computations. We examine the effects of false positives and negatives, blur, dataset biases, and spatial deviations on performance.
This paper offers a more complete understanding of evaluation metrics and what they measure. 

\subsection{Qualitative evaluation of saliency}
\label{sec:qualitative_eval}

Most saliency papers include side-by-side comparisons of different saliency maps computed for the same images (as in Fig.~\ref{fig:main}b). Visualizations of saliency maps are often used to highlight improvements over previous models.
A few anecdotal images might be used to showcase model strengths and weaknesses. 

Bruce et al.~\cite{bruce2015computational} discussed the problems with visualizing saliency maps, in particular the strong effect that contrast has on the perception of saliency models. 
We propose supplementing saliency map examples with visualizations of metric computations (as in Fig.~\ref{fig:vis} and throughout the rest of this paper) to provide an additional means of comparison that is more tightly linked to the underlying model performance than the saliency maps themselves.


\section{Evaluation setup}
\label{sec:evaluationSetup}

The choice of evaluation metrics should be considered in the context of the whole evaluation setup, which requires the following decisions to be made:
(1)~on which input images saliency models will be evaluated, (2)~how the ground truth eye movements will be collected (e.g. at which distance and for how long human observers view each image), and (3)~how the eye movements will be represented (e.g. as discrete points, sequences, or distributions). 
In this section we explain the design choices used for our data collection and evaluation.

\subsection{Data collection} 


We used the MIT Saliency Benchmark dataset (\textbf{MIT300}) of 300 natural images~\cite{mit-saliency-benchmark,Judd_2012}. 
Eye movements were collected by allowing participants to free-view each image for 2 seconds (more details in the appendix). 
Such a viewing duration typically elicits 4-6 fixations from each observer. This is sufficient to highlight a few points of interest per image, and offers a reasonable testing ground for saliency models.  
Different tasks (free viewing, visual search, etc.) also differently direct eye movements and may require alternative model assumptions~\cite{BylinskiiOpinion}. The free viewing task is most commonly used for saliency modeling as it requires fewest additional assumptions. 


The eye tracking set-up, including participant distance to the eye tracker, calibration error, and image size affects the assumptions that can be made about the collected data. 
In the eye tracking set-up of the MIT300 dataset, one degree of visual angle is approximately 35 pixels. One degree of visual angle is typically used both (1) as an estimate of the size of the human fovea: e.g. how much of the image a participant has in focus during a fixation, and (2) to account for measurement error in the eye tracking set-up.
The robustness of the data also depends on the number of eye fixations collected. 
In the MIT300 dataset, the eye fixations of 39 observers are available per image, more than in other datasets of similar size.

\subsection{Ground truth representation}
\label{sec:groundtruthrep}

Once collected, the ground truth eye fixations can be processed and formatted in a number of ways for saliency evaluation.
There is a fundamental ambiguity in the correct representation for the fixation data, and different representational choices rely on different assumptions.
One format is to use the original fixation locations.
Alternatively, the discrete fixations can be converted into a continuous distribution, a \textbf{fixation map}, by smoothing (Fig.~\ref{fig:main}a). We follow common practice\footnote{Some researchers choose to cross-validate the smoothing parameter instead of fixing it as a function of viewing angle \cite{kummerer2014close,kummerer2015information}.} and blur each fixation location using a Gaussian with sigma equal to one degree of visual angle~\cite{LeMeur2013}. In the following section, we denote the map of fixation locations as $Q^{B}$  and the continuous fixation map (distribution) as $Q^{D}$.

Smoothing the fixation locations into a continuous map acts as regularization. 
It allows for uncertainty in the ground truth measurements to be incorporated: error in the eye-tracking as well as uncertainty of what an observer sees when looking at a particular location on the screen. Any splitting of observer fixations in two sets will never lead to perfect overlap (due to the discrete nature of the data), and smoothing provides additional robustness for evaluation. In the case of few observers, smoothing the fixation locations helps to extrapolate the existing data. 

On the other hand, conversion of the fixation locations into a distribution requires parameter selection and post-processing the collected data. The smoothing parameter can significantly affect metric scores during evaluation (Table~\ref{fig:synthetic1}a), unless the model itself is properly regularized.

The fixation locations can be viewed as a discrete sample from some ground truth distribution that the fixation map attempts to approximate. Similarly, the fixation map can be viewed as an extrapolation of discrete fixation data to the case of infinite observers. 


Metrics for the evaluation of sequences of fixations are also available~\cite{LeMeur2013}. However, most saliency models and evaluations are tuned for location prediction, as sequences tend to be noisier and harder to evaluate. We only consider spatial, not temporal, fixation data.

\begin{figure*}
\centering
\includegraphics[width=0.9\linewidth]{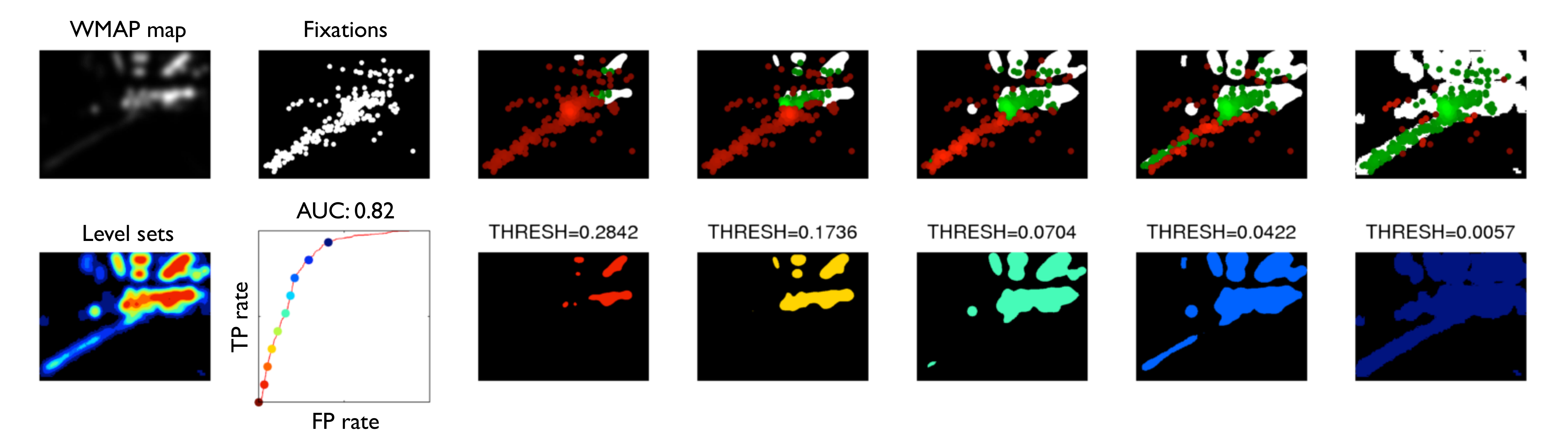}
\caption{{\small The AUC metric evaluates a saliency map's predictive power by how many ground truth fixations it captures in successive level sets. To compute AUC, a saliency map (top left) is treated as a binary classifier of fixations at various threshold values (THRESH) and an ROC curve is swept out. Thresholding the saliency map produces the level sets in the bottom row. For each level set, the true positive rate is the proportion of fixations landing in the level set (top row, green points). The false positive rate is the proportion of image pixels in the level set not covered in fixations. We include 5 level sets corresponding to points on the ROC curve. The AUC score for the saliency map is the area under the ROC curve.}}
\label{fig:auc_curve}
\vspace{-1em}
\end{figure*}

\section{Metric computation}
\label{sec:metrics}

In this paper, we study saliency metrics, that is, functions that take two inputs representing eye fixations (ground truth and predicted) and then output
a number assessing the similarity or dissimilarity between them. Given a set of ground truth eye fixations, such metrics can be used to define scoring functions, which take a saliency map prediction as input and return a number assessing the accuracy of the prediction. The definition of a score can further involve post-processing (or regularizing) the prediction to conform it to known characteristics of the ground truth and ignore potentially distracting idiosyncratic errors. In this paper, we focus on the metric and not on the regularization of ground truth data.

We consider 8 popular saliency evaluation metrics in their most common variants. 
Some metrics have been designed specifically for saliency evaluation (shuffled AUC, Information Gain, and Normalized Scanpath Saliency), while others have been adapted from signal detection (variants of AUC), image matching and retrieval (Similarity, Earth Mover's Distance), information theory (KL-divergence), and statistics (Pearson's Correlation Coefficient).
Because of their original intended applications, these metrics expect different input formats: KL-divergence and Information Gain expect valid probability distributions as input, Similarity and Earth Mover's Distance can operate on unnormalized densities and histograms, while Pearson's Correlation Coefficient (CC) treats its inputs as random variables.


%
One of the intentions of this paper is to serve as a guide to complement the MIT Saliency Benchmark, and to provide interpretation for metric scores. The MIT Saliency Benchmark accepts saliency maps as intensity maps, without restricting input to be in any particular form (probabilistic or otherwise). 
If a metric expects valid probability distributions, we normalize the input saliency maps accordingly, but otherwise make no additional modifications or optimizations. 

In this paper we analyze these 8 metrics in isolation from the input format and with minimal underlying assumptions. The only distinction we make in terms of the input that these metrics operate on is whether the ground-truth is represented as discrete fixation locations or a continuous fixation map. Accordingly, we categorize metrics as \textbf{location-based} or \textbf{distribution-based} (following Riche et al.~\cite{RicheICCV2013}). This organization is summarized in Table~\ref{tab:metric_summary}. In this section, we discuss the particular advantages and disadvantages of each metric, and present visualizations of the metric computations. Additional variants and implementation details are provided in the appendix. 





\begin{table}
\centering
\begin{tabular}{| p{1.6cm} | p{2.7cm} | p{2.5cm} | }
\hline
\textbf{Metrics} & \textbf{Location-based} & \textbf{Distribution-based}  \\ \hline
\textbf{Similarity} & AUC, sAUC, NSS, IG & SIM, CC \\ \hline
\textbf{Dissimilarity}  & & EMD, KL\\ \hline
\end{tabular}
\caption{{\small Different metrics use different formats of ground truth for evaluating saliency models. Location-based metrics consider saliency map values at discrete fixation locations, while distribution-based metrics treat both ground truth fixation maps and saliency maps as continuous distributions. Good saliency models should have high values for similarity metrics and low values for dissimilarity metrics.}}
\vspace{-1.5em}
\label{tab:metric_summary}
\end{table}

\subsection{Location-based metrics}

\subsubsection{Area under ROC Curve (AUC): \\Evaluating saliency as a classifier of fixations} \label{sec:AUC}


Given the goal of predicting the fixation locations on an image, a saliency map can be interpreted as a classifier of which pixels are fixated or not. This suggests a detection metric for measuring saliency map performance.
In signal detection theory, the Receiver Operating Characteristic (ROC) measures the tradeoff between true and false positives at various discrimination thresholds  \cite{Green1966,fawcett2006introduction}.
The Area under the ROC curve, referred to as AUC, is the most widely used metric for evaluating saliency maps. The saliency map is treated as a binary classifier of fixations at various threshold values (level sets), and an ROC curve is swept out by measuring the true and false positive rates under each binary classifier (level set). 
Different AUC implementations differ in how true and false positives are calculated. 
Another way to think of AUC is as a measure of how well a model performs on a 2AFC task, where given 2 possible locations on the image, the model has to pick the location that corresponds to a fixation~\cite{kummerer2015information}.\\


\noindent\textbf{Computing true and false positives:}

An AUC variant from Judd et al. \cite{Judd_2009}, called \textbf{AUC-Judd} \cite{mit-saliency-benchmark}, is depicted in Fig.~\ref{fig:auc_curve}. For a given threshold, the true positive rate (\textbf{TP rate}) is the ratio of true positives to the total number of fixations, where true positives are saliency map values above threshold at \emph{fixated pixels}. This is equivalent to the ratio of fixations falling within the level set to the total fixations. %

The false positive rate (\textbf{FP rate}) is the ratio of false positives to the total number of saliency map pixels 
at a given threshold, where false positives are saliency map values above threshold at \emph{unfixated pixels}. 
This is equivalent to the number of pixels in each level set, minus the pixels already accounted for by fixations.

\begin{figure*}
\centering
\includegraphics[width=0.8\linewidth]{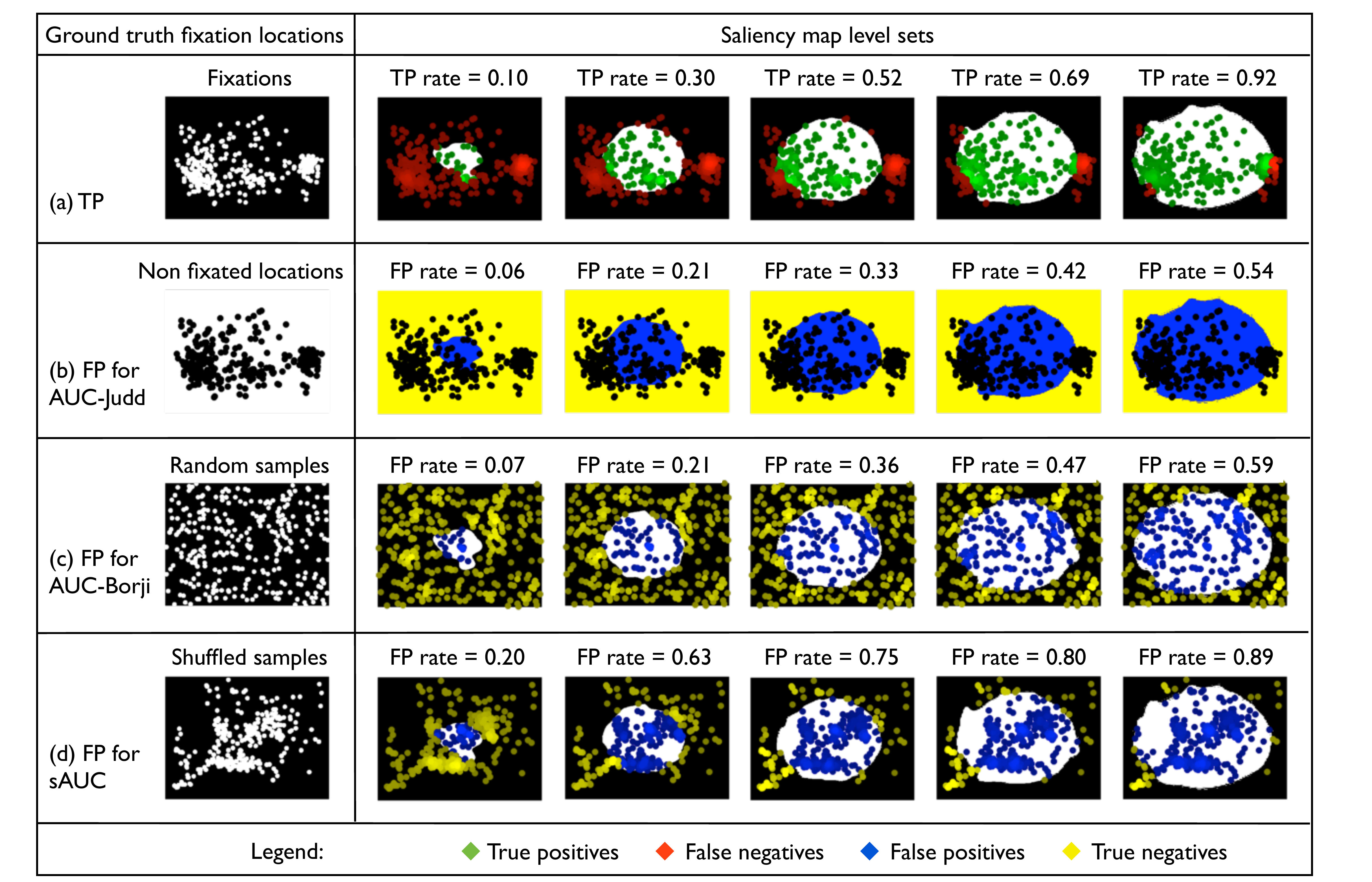}
\caption{{\small How true and false positives are calculated under different AUC metrics: (a) In all cases, the true positive rate is calculated as the proportion of fixations falling into the thresholded saliency map (green over green plus red). (b) In AUC-Judd, the false positive rate is the proportion of non-fixated pixels in the thresholded saliency map (blue over blue plus yellow). (c) In AUC-Borji, this calculation is approximated by sampling negatives uniformly at random and computing the proportion of negatives in the thresholded region (blue over blue plus yellow). (d) In sAUC, negatives are sampled according to the distribution of fixations in other images instead of uniformly at random. Saliency models are scored similarly under the AUC-Judd and AUC-Borji metrics, but differently under sAUC due to the sampling of false positives.}}
\label{fig:labeling_strats}
\vspace{-1em}
\end{figure*}

Another variant of AUC by Borji et al. \cite{Borji_2012}, called \textbf{AUC-Borji} \cite{mit-saliency-benchmark}, uses a uniform random sample of image pixels as negatives 
and defines the saliency map values above threshold at these pixels as false positives. These AUC implementations are compared in Fig.~\ref{fig:labeling_strats}. The first row depicts the TP rate calculation, equivalent across implementations. The second and third rows depict the FP rate calculations in AUC-Judd and AUC-Borji, respectively. The false positive calculation in AUC-Borji is a discrete approximation of the calculation in AUC-Judd. 
Because of a few approximations in the AUC-Borji implementation that can lead to suboptimal behavior, we report AUC scores using AUC-Judd in the rest of the paper.
Additional discussion, implementation details, and other variants of AUC are discussed in the appendix.\\

\noindent\textbf{Penalizing models for center bias:}

The natural distribution of fixations on an image tends to include a higher density near the center of an image~\cite{Tatler2007JOV}. 
As a result, a model that incorporates a center bias into its predictions will be able to account for at least part of the fixations on an image, independent of image content. In a center-biased dataset, a center prior baseline will achieve a high AUC score. 

The shuffled AUC metric, \textbf{sAUC}~\cite{BorjiICCV2013,Tatler2007JOV,Zhang2008JOV,Einhauser2003,Tatler2005} samples negatives from fixation locations from other images, instead of uniformly at random. This has the effect of sampling negatives predominantly from the image center because averaging fixations over many images results in the natural emergence of a central Gaussian distribution~\cite{Tatler2007JOV,wilming2011measures}. 
In Fig.~\ref{fig:labeling_strats} the shuffled sampling strategy of sAUC is compared to the random sampling strategy of AUC-Borji. 

\begin{figure}
\centering
\includegraphics[width=0.9\linewidth]{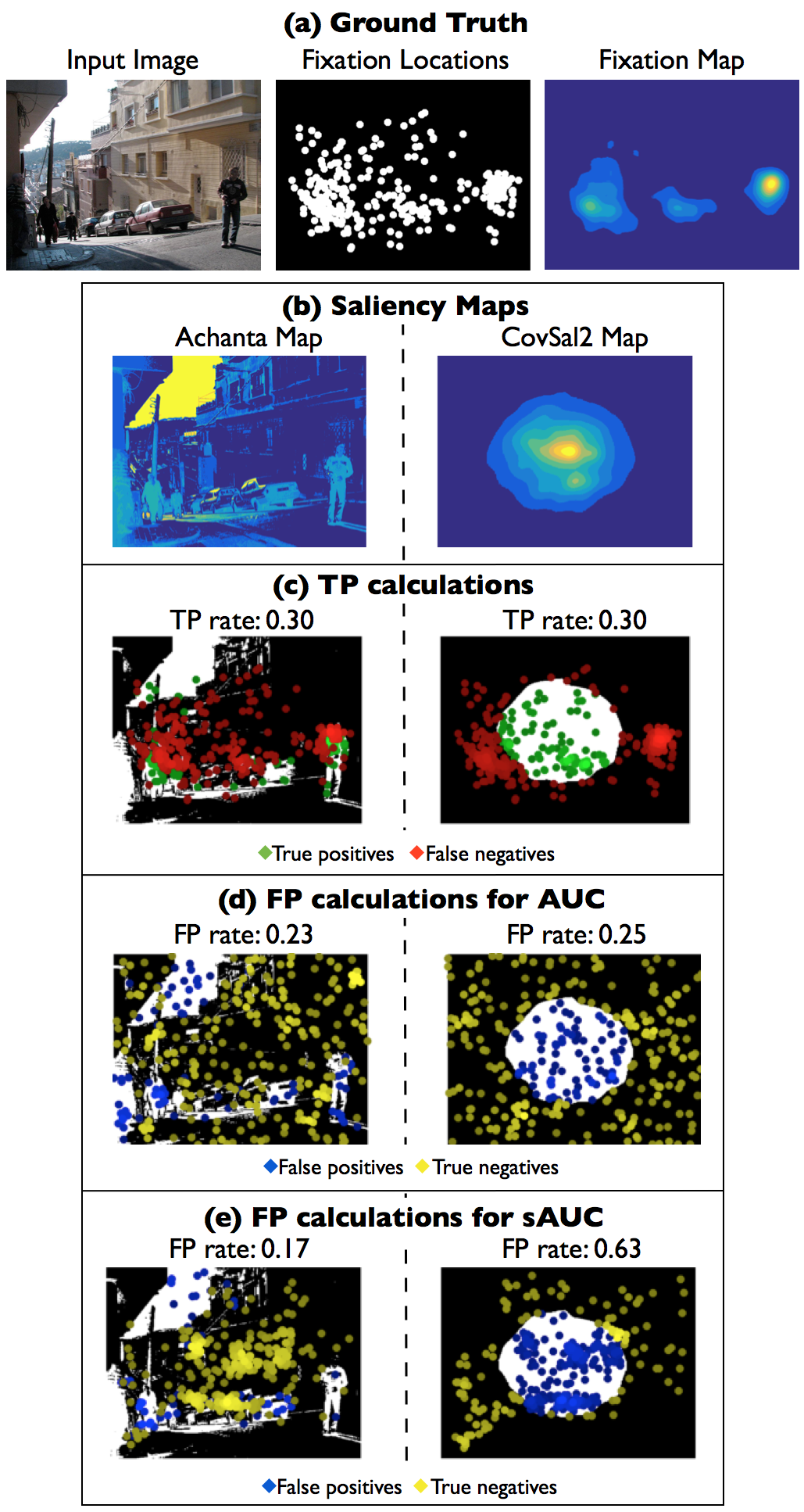}
\caption{{\small Both AUC and sAUC measure the ability of a saliency map to classify fixated from non-fixated locations (Sec.~\ref{sec:AUC}). The main difference is that AUC prefers maps that account for center bias, while sAUC penalizes them.
The saliency maps in (b) are compared on their ability to predict the ground truth fixations in (a). 
For a particular level set, the true positive rate is the same for both maps (c). The sAUC metric normalizes this value by fixations sampled from other images, more of which land in the center of the image, thus penalizing the rightmost model for its center bias (d). The AUC metric, however, samples fixations uniformly at random and prefers the center-biased model which better explains the overall viewing behavior~(e).}}
\label{fig:centerbias}
\vspace{-2em}
\end{figure}

A model that only predicts the center achieves an sAUC score of 0.5 because at all thresholds this model captures as many fixations on the target image as on other images (TP rate = FP rate). A model that incorporates a center bias into its predictions is putting density in the center at the expense of other image regions. Such a model will score worse according to sAUC compared to a model that makes off-center predictions, because sAUC will effectively discount the central predictions (Fig.~\ref{fig:centerbias}). In other words, sAUC is not invariant to whether the center bias is modeled: it specifically penalizes models that include the center bias.\\


\begin{figure}
\centering
\includegraphics[width=1\linewidth]{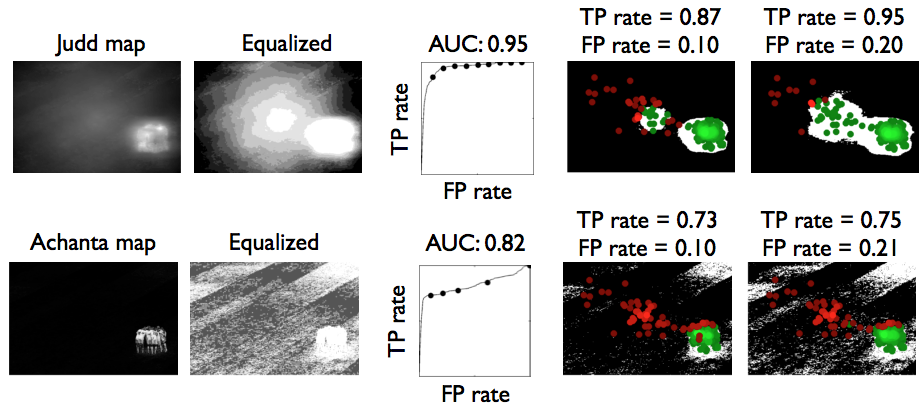}
\caption{{\small The saliency map in the top row accounts for more fixations in its first few level sets than the map in the bottom row, achieving a higher AUC score overall. The AUC score is driven most by the first few level sets, while the total number of levels sets and false positives in later level sets have a significantly smaller impact. Equalizing the saliency map distributions allows us to visualize the level sets. The map in the bottom row has a smaller range of saliency values, and thus fewer level sets and sample points on the ROC curve. Both axes on the ROC curves span 0 to 1.}}
\label{fig:first_level_sets}
\vspace{-1em}
\end{figure}

\noindent\textbf{Invariance to monotonic transformations:}

AUC metrics measure only the relative (i.e., ordered) saliency map values at ground truth fixation locations. In other words, the AUC metrics are ambivalent to monotonic transformations. 
AUC is computed by varying the threshold of the saliency map and computing a trade-off between true and false positives.
Lower thresholds correspond to measuring the coverage similarity between distributions, while higher thresholds correspond to measuring the similarity between the peaks of the two maps \cite{engelke2013comparative}.
Due to how the ROC curve is computed, the AUC score for a saliency map is mostly driven by the higher thresholds: i.e.,~the number of ground truth fixations captured by the peaks of the saliency map (or the first few level sets as in Fig.~\ref{fig:first_level_sets}). 
Models that place high-valued predictions at fixated locations receive high scores, while low-valued predictions at non-fixated locations are mostly ignored (Sec.~\ref{sec:misses_falsealarms}). 

\subsubsection{Normalized Scanpath Saliency (NSS): \\Measuring the normalized saliency at fixations} \label{sec:NSS}

The Normalized Scanpath Saliency, \textbf{NSS} was introduced to the saliency community as a simple correspondence measure between saliency maps and ground truth, computed as the average normalized saliency at fixated locations \cite{Peters2005}.
Unlike in AUC, the absolute saliency values are part of the normalization calculation. 
NSS is sensitive to false positives, relative differences in saliency across the image, and general monotonic transformations. 
However, because the mean saliency value is subtracted during computation, NSS is invariant to linear transformations like contrast offsets.  
Given a saliency map $P$ and a binary map of fixation locations $Q^{B}$:

\small
\begin{equation} \label{eq:NSS}
\begin{split}
NSS(P,Q^{B}) = \frac{1}{N}\sum_{i} \overline{P_i} \times Q^{B}_i \\
\mbox{ where } N = \sum_{i}Q^{B}_i\mbox{ and } \overline{P} = \frac{P-\mu(P)}{\sigma(P)}
\end{split}
\end{equation}
\normalsize

\noindent where $i$ indexes the $i^{th}$ pixel, and $N$ is the total number of fixated pixels.
Chance is at 0, positive NSS indicates correspondence between maps above chance, and negative NSS indicates anti-correspondence. For instance, a unity score corresponds to fixations falling on portions of the saliency map with a saliency value one standard deviation above average.

\begin{figure}
\centering
\includegraphics[width=1\linewidth]{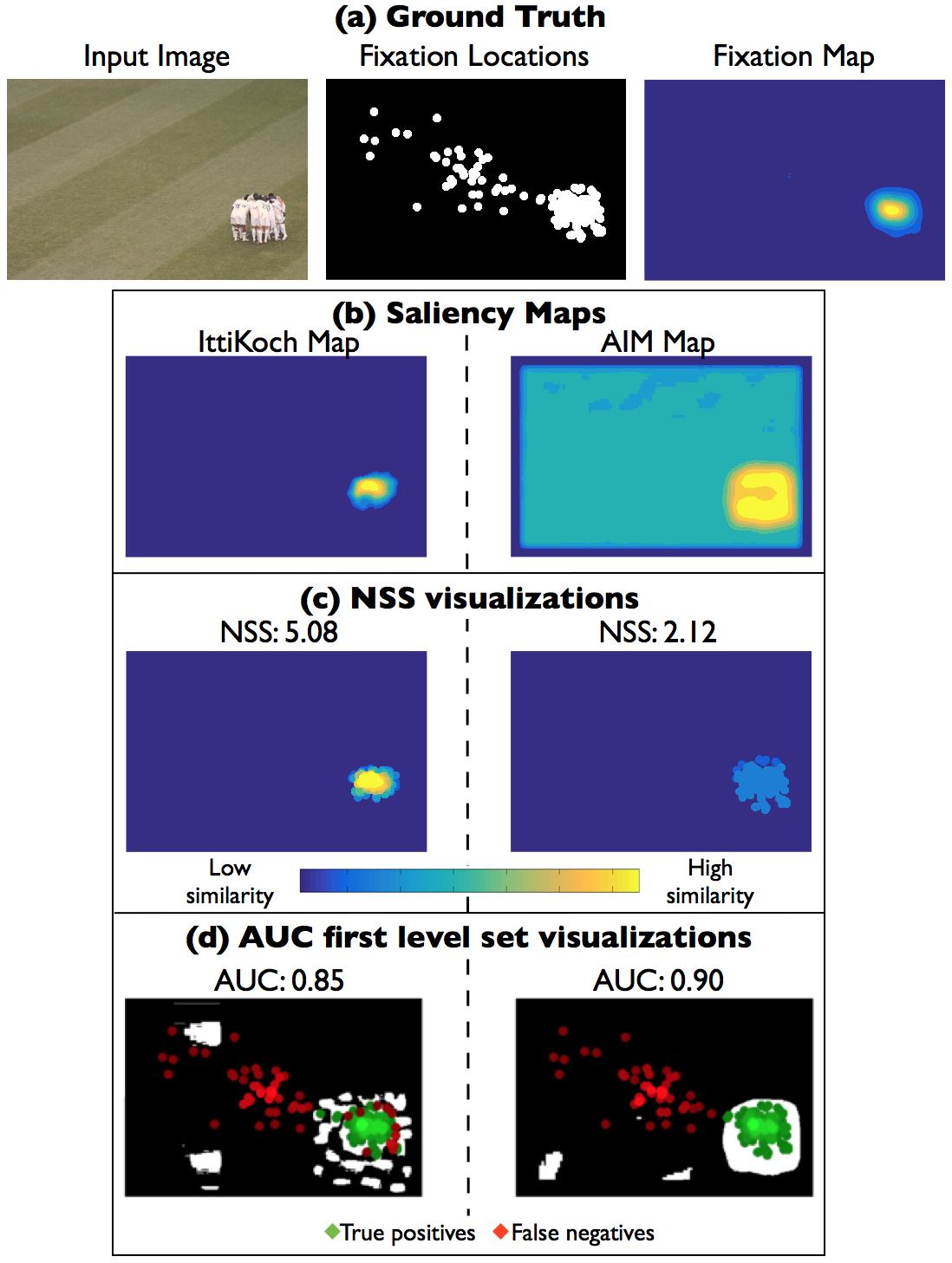}
\caption{{\small Both AUC and NSS evaluate the ability of a saliency map (b) to predict fixation locations (a). AUC is invariant to monotonic transformations (Sec.~\ref{sec:AUC}), while NSS is not. NSS normalizes a saliency map by the standard deviation of the saliency values (Sec.~\ref{sec:NSS}). 
AUC ignores low-valued false positives but NSS penalizes them. 
As a result, the rightmost map has a lower NSS score because more false positives means the normalized saliency value at fixation locations drops (c). The AUC score of the left and right maps is very similar since a similar number of fixations fall in equally-sized level sets of the two saliency maps (d). 
}}
\label{fig:AUC_vs_NSS}
\vspace{-1em}
\end{figure}

Recall that a saliency model with high-valued predictions at fixated locations would receive a high AUC score even in the presence of many low-valued false positives (Fig.~\ref{fig:AUC_vs_NSS}d). However, all false positives contribute to lowering the normalized saliency value at each fixation location, thus reducing the overall NSS score (Fig.~\ref{fig:AUC_vs_NSS}c). The visualization for NSS consists of the normalized saliency value for each fixation location (i.e., $\overline{P_i}$ where $Q^{B}_i = 1$).

\subsubsection{Information Gain (IG): \\Evaluating information gain over a baseline}

Information Gain, \textbf{IG}, was recently introduced by K{\"u}mmerer et al. \cite{kummerer2014close,kummerer2015information} as an information theoretic metric that measures saliency model performance beyond systematic bias (e.g., a center prior baseline). 

Given a binary map of fixations $Q^{B}$, a saliency map $P$, and a baseline map $B$, information gain is computed as:

\small
\begin{equation} \label{eq:IG}
IG(P,Q^{B}) = \frac{1}{N}\sum_{i}Q^{B}_{i}[\log_2(\epsilon + P_{i}) - \log_2(\epsilon + B_{i})] 
\end{equation}
\normalsize

\noindent where $i$ indexes the $i^{th}$ pixel, $N$ is the total number of fixated pixels, $\epsilon$ is for regularization, and information gain is measured in bits per fixation. This metric measures the average information gain of the saliency map over the center prior baseline at fixated locations (i.e., where $Q^{B}=1$). 

IG assumes that the input saliency maps are probabilistic, properly regularized and optimized to include a center prior~\cite{kummerer2014close,kummerer2015information}. 
A score above zero indicates the saliency map predicts the fixated locations better than the center prior baseline. This score measures how much image-specific saliency is predicted beyond image-independent dataset biases, which in turn requires careful modeling of these biases.

\begin{figure}
\includegraphics[width=1\linewidth]{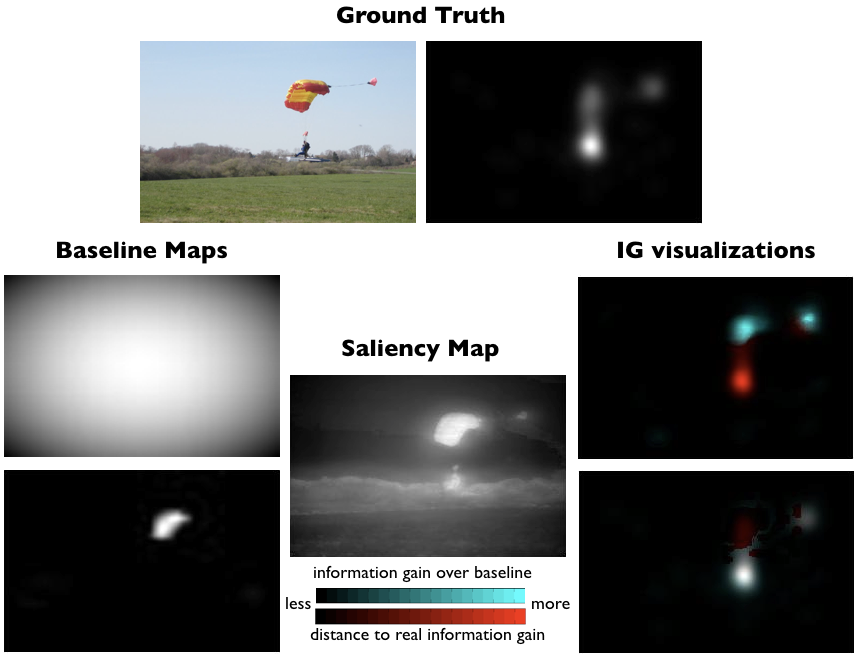}
\caption{{\small We compute the information of one model over another at predicting ground truth fixations. We visualize the information gain of the Judd model over the center prior baseline (top) and the bottom-up IttiKoch model (bottom). In blue are the image pixels where the Judd model makes better predictions than each model. In red is the remaining distance to the real information gain: i.e., image pixels at which the Judd model underestimates saliency.}}
\label{fig:IG_visualization}
\vspace{-2em}
\end{figure}

We can also compute the information gain of one model over another to measure how much image-specific saliency is captured by one model beyond what is already captured by another model.
The example in Fig.~\ref{fig:IG_visualization} contains a visualization of the information gain of the Judd model over the center prior baseline and over the bottom-up IttiKoch model. Visualized in red are image regions for which the Judd model underestimates saliency relative to each model, and in blue are image regions for which the Judd model achieves a gain in performance over each model at predicting the ground truth. The human under the parachute has a high saliency under the center prior model, while the Judd model underestimates the relative saliency of this area (red), but the parachute is where the Judd model has positive information gain over the center prior (blue). On the other hand, the bottom-up IttiKoch model captures the parachute but misses the person in the center of the image, so in this case the Judd model achieves gains on the central image pixels but not on the parachute. We refer the reader to \cite{kummerer2015information} for a more detailed discussion and visualizations of the IG metric.

\subsection{Distribution-based metrics}

The (location-based) metrics described so far score saliency models at how accurately they predict discrete fixation locations. If the ground truth fixation locations are interpreted as a possible sample from some underlying probability distribution, then another approach is to predict the underlying distribution directly instead of the fixation locations.
Although we can not directly observe the ground truth distribution, it is often approximated by Gaussian blurring the fixation locations into a fixation map (Sec.~\ref{sec:groundtruthrep}). In this next section we discuss a set of metrics that score saliency models at how accurately they approximate the continuous fixation map.

\subsubsection{Similarity (SIM): \\Measuring the intersection between distributions}
\label{sec:SIM}

\begin{figure}
\centering
\includegraphics[width=0.9\linewidth]{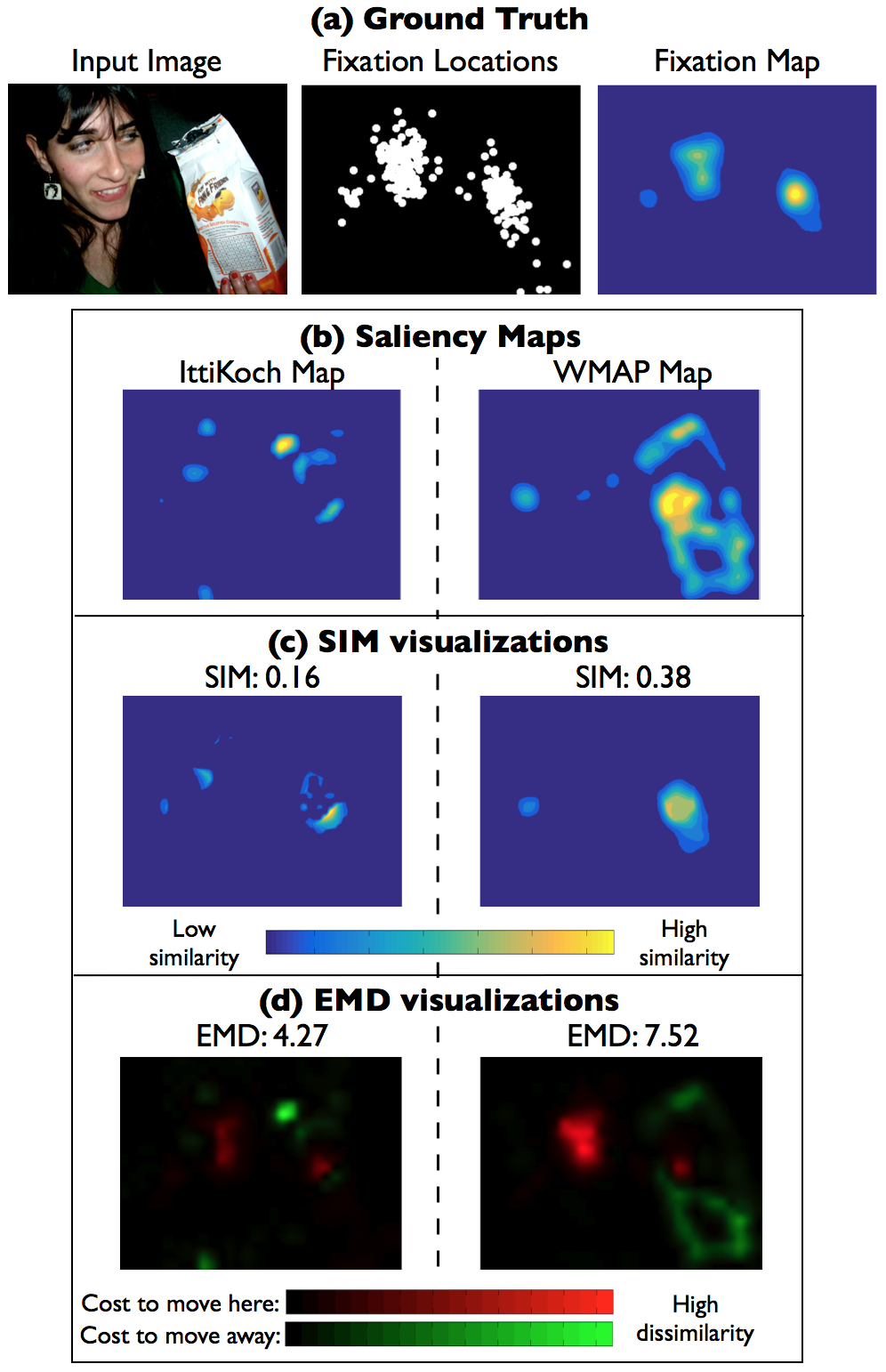}
\caption{{\small The EMD and SIM metrics measure the similarity between the saliency map (b) and ground truth fixation map (a). EMD measures how much density needs to be moved before the two maps match (Sec.~\ref{sec:EMD}), while SIM measures the direct intersection between two maps (Sec.~\ref{sec:SIM}).
EMD prefers sparser predictions, even if they do not perfectly align with fixated regions, while SIM penalizes misalignment. 
The saliency map on the left makes sparser predictions, resulting in a smaller intersection with the ground truth, and lower SIM score, than the map on the right (c). The predicted density in the leftmost map is spatially closer to the ground truth density than the density in the rightmost map, and achieves a better EMD score (d). }}
\label{fig:S_vs_EMD}
\vspace{-1em}
\end{figure}

The similarity metric, \textbf{SIM} (also referred to as \emph{histogram intersection}), measures the similarity between two distributions, viewed as histograms. First introduced as a metric for color-based and content-based image matching \cite{Rubner2000,swain1991color}, it has gained popularity in the saliency community as a simple comparison between pairs of saliency maps. 
SIM is computed as the sum of the minimum values at each pixel, after normalizing the input maps.
Given a saliency map $P$ and a continuous fixation map $Q^{D}$:

\small
\begin{equation} \label{eq:SIM}
\begin{split}
SIM(P,Q^{D}) = \sum_{i} \min(P_{i}, Q^{D}_{i}) \\
\mbox{ where } \sum_{i}P_{i} = \sum_{i}Q^{D}_{i} =1
\end{split}
\end{equation}
\normalsize

\noindent iterating over discrete pixel locations $i$. 
A SIM of one indicates the distributions are the same, while a SIM of zero indicates no overlap. 
Fig.~\ref{fig:S_vs_EMD}c contains a visualization of this operation. At each pixel $i$ of the visualization, we plot $\min(P_{i}, Q^{D}_{i})$.
Note that the model with the sparser saliency map has a lower histogram intersection with the ground truth map. SIM is very sensitive to missing values, and penalizes predictions that fail to account for all of the ground truth density (see Sec.~\ref{sec:misses_falsealarms} for a discussion). \\

\noindent\textbf{Effect of blur on model performance:}

The downside of a distribution metric like SIM is that the choice of the Gaussian sigma (or blur) in constructing the fixation and saliency maps affects model evaluation. For instance, as demonstrated in the synthetic experiment in Fig.~\ref{fig:synthetic1}a, even if the correct location is predicted, SIM will only reach its maximal value when the saliency map's sigma exactly matches the ground truth sigma. The SIM score drops off drastically under different sigma values, more than the other metrics. 
Fine-tuning this blur value on a training set with similar parameters as the test set (eyetracking set-up, viewing angle) can help boost model performances \cite{mit-saliency-benchmark,Judd_2012}.

The SIM metric is good for evaluating partial matches, where a subset of the saliency map accounts for the ground truth fixation map. As a side-effect, false positives tend to be penalized less than false negatives. For other applications, a metric that treats false positives and false negatives symmetrically, such as CC or NSS, may be preferred.

\subsubsection{Pearson's Correlation Coefficient (CC): \\Evaluating the linear relationship between distributions} \label{sec:CC}

\begin{figure}
\includegraphics[width=0.9\linewidth]{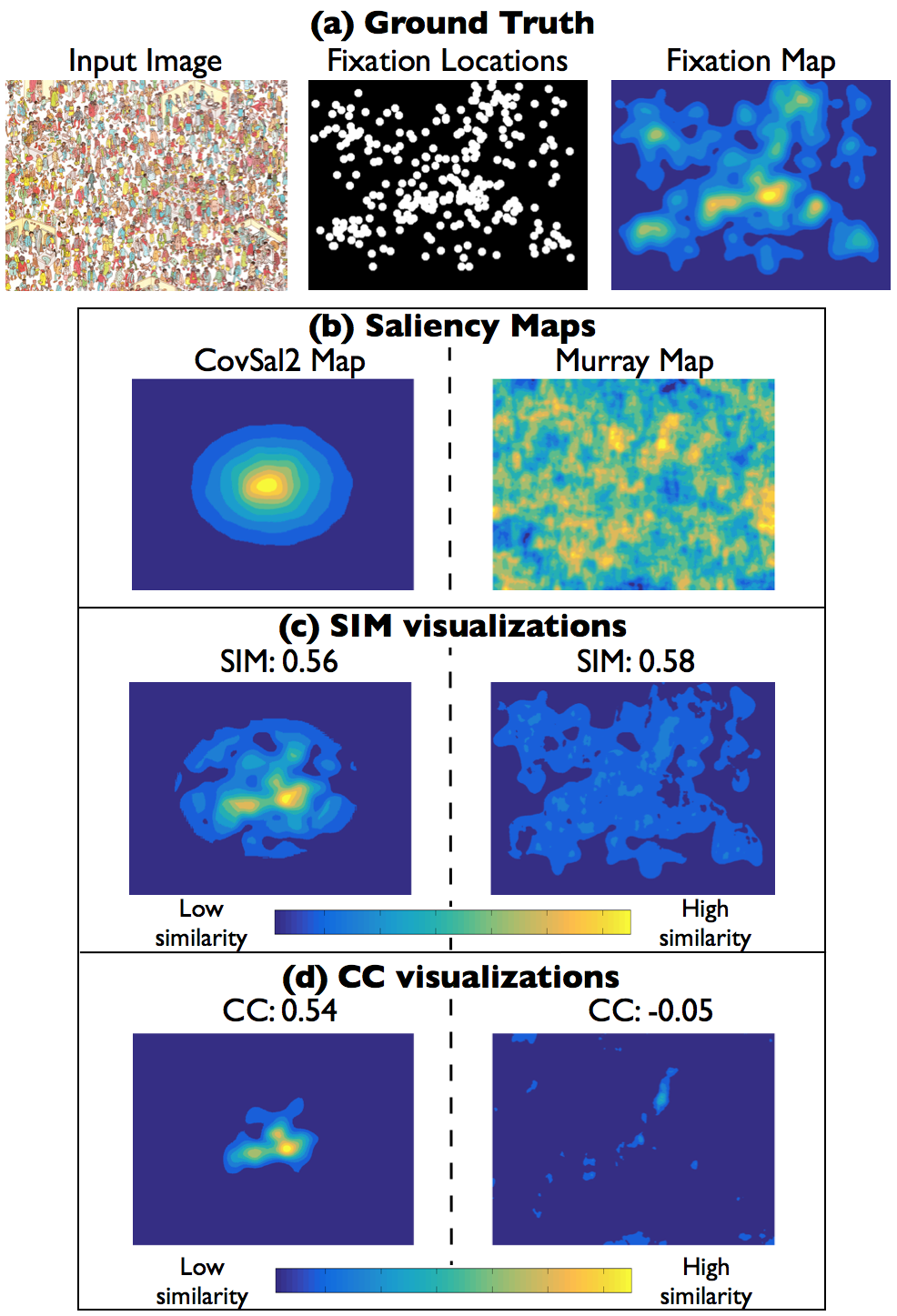}
\caption{{\small The SIM and CC metrics measure the similarity between the saliency map (b) and ground truth fixation map (a). SIM measures the histogram intersection between two maps (Sec.~\ref{sec:SIM}), while CC measures their cross correlation (Sec.~\ref{sec:CC}). CC treats false positives and negatives symmetrically, but SIM places less emphasis on false positives than false negatives. As a result,
both saliency maps have similar SIM scores (c), but the saliency map on the right has a lower CC score because false positives lower the overall correlation (d). }} 
\label{fig:S_vs_CC}
\vspace{-1.8em}
\end{figure}

The Pearson's Correlation Coefficient, \textbf{CC}, also called \emph{linear correlation coefficient} is a statistical method used generally in the sciences for measuring how correlated or dependent two variables are. 
CC can be used to interpret saliency and fixation maps, $P$ and $Q^{D}$, as random variables to measure the linear relationship between them \cite{LeMeur_VR2007}: 

\small
\begin{equation} \label{eq:CC}
CC(P,Q^{D}) =  \frac{\sigma(P,Q^{D})}{\sigma(P)\times\sigma(Q^{D})}
\end{equation}
\normalsize

\noindent where $\sigma(P,Q^{D})$ is the covariance of $P$ and $Q^{D}$. 
CC~is symmetric and penalizes false positives and negatives equally. It is invariant to linear (but not arbitrary monotonic) transformations. 
High positive CC values occur at locations where both the saliency map and ground truth fixation map have values of similar magnitudes. 
Fig.~\ref{fig:S_vs_CC} is an illustrative example comparing the behaviors of SIM and CC: where SIM penalizes false negatives significantly more than false positives, but CC treats both symmetrically. 
For visualizing CC in Fig.~\ref{fig:S_vs_CC}d, each pixel $i$ has value: 

\small
\begin{equation} \label{eq:CCvis}
V_i = \frac{P_i\times Q^{D}_i}{\sqrt{\sum_{j}(P_j^2 + (Q^{D}_j)^2)} }
\end{equation}
\normalsize

\noindent Due to its symmetric computation, CC can not distinguish whether differences between maps are due to false positives or false negatives. Other metrics may be preferable if this kind of analysis is of interest.

\subsubsection{Kullback-Leibler divergence (KL):\\ Evaluating saliency with a probabilistic interpretation} \label{sec:KL}

\begin{figure}
\includegraphics[width=0.9\linewidth]{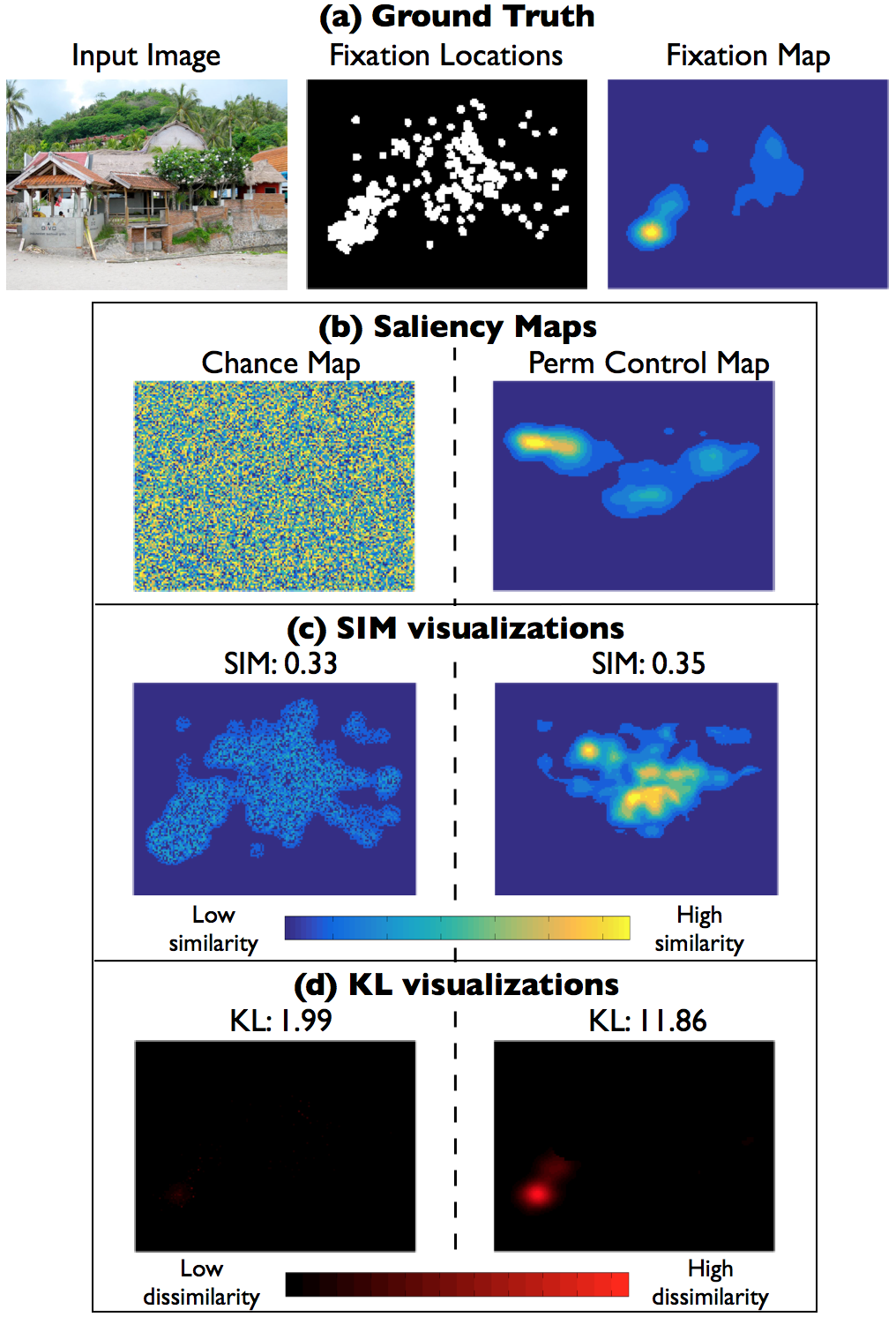}
\caption{{\small The SIM and KL metrics measure the similarity between the saliency map (b) and ground truth fixation map (a), treating the former as the predicted distribution and the latter as the target distribution. SIM measures the histogram intersection between the distributions (Sec.~\ref{sec:SIM}), while KL measures an information-theoretic divergence between the two distributions (Sec.~\ref{sec:KL}). KL is much more sensitive to false negatives that SIM. 
Both saliency maps in (b) are image-agnostic baselines. They receive similar scores under the SIM metric (c). However, because the map on the left places uniformly-sampled saliency values at all image pixels, it contains fewer zero values, and is favored by KL (d). The rightmost map samples saliency from another image, resulting in zero-values at multiple fixated locations, and a poor KL score (d).}}
\label{fig:S_vs_KL}
\vspace{-2.5em}
\end{figure}

Kullback-Leibler (\textbf{KL}) is a general information theoretic measure of the difference between two probability distributions. In the saliency literature, depending on how the saliency predictions and ground truth fixations are interpreted as distributions, different KL computations are possible. We discuss a few alternative varieties in the appendix. To avoid future confusion about the KL implementation used, we can refer to this variant as \textbf{KL-Judd} similarly to how the AUC variant traditionally used on the MIT Benchmark is referred to as AUC-Judd.
Analogous to our other distribution-based metrics, our KL metric takes as input a saliency map $P$ and a ground truth fixation map $Q^{D}$, and
evaluates the loss of information when $P$ is used to approximate $Q^{D}$:

\small
\begin{equation} \label{eq:KL}
KL(P,Q^{D}) = \sum_{i}Q^{D}_{i} \log\left(\epsilon + \frac{Q^{D}_{i}}{\epsilon+P_{i}}\right)
\end{equation}
\normalsize

\noindent where $\epsilon$ is a regularization constant\footnote{The relative magnitude of $\epsilon$ will affect the regularization of the saliency maps and how much zero-valued predictions are penalized. The MIT Saliency Benchmark uses MATLAB's built-in \emph{eps} with value = 2.2204e-16.}.
KL-Judd is an asymmetric dissimilarity metric, with a lower score indicating a better approximation of the ground truth by the saliency map. 
We compute a per-pixel score to visualize the KL computation (Fig.~\ref{fig:S_vs_KL}d). For each pixel $i$ in the visualization, we plot $Q^{D}_{i} \log\left(\epsilon + \frac{Q^{D}_{i}}{\epsilon+P_{i}}\right)$. Wherever the ground truth value $Q^{D}_i$ is non-zero but $P_i$ is close to or equal to zero, a large quantity is added to the KL score. Such regions are the brightest in the KL visualization. There are more bright regions in the rightmost map of Fig.~\ref{fig:S_vs_KL}d, corresponding to areas in the ground truth map that were left unaccounted for by the predicted saliency. 
Both models compared in Fig.~\ref{fig:S_vs_KL} are image-agnostic: one is a chance model that assigns a uniform value to each pixel in the image, and the other is a \textbf{permutation control} model which uses a fixation map from another randomly-selected image. The permutation control model is more likely to capture viewing biases common across images. It scores above chance for many of the metrics in Table~\ref{tab:centerperf}. However, KL is so sensitive to zero-values that a sparse set of predictions is penalized very harshly, significantly worse than chance.

\begin{table*}
\centering
\begin{tabular}{| p{0.15\linewidth} | p{0.07\linewidth}  | p{0.07\linewidth} | p{0.07\linewidth} | p{0.07\linewidth} | p{0.07\linewidth}| p{0.07\linewidth} | p{0.07\linewidth} | p{0.07\linewidth} |}
\hline
Saliency model &\multicolumn{6}{c |}{Similarity metrics}&\multicolumn{2}{c |}{Dissimilarity metrics}\\
\hline
 & SIM $\uparrow$ & CC $\uparrow$ & NSS $\uparrow$ & AUC $\uparrow$ & sAUC $\uparrow$ & IG $\uparrow$ & KL $\downarrow$ & EMD $\downarrow$ \\ \hline
Infinite Observers & 1.00 & 1.00 & 3.29 & 0.92 & 0.81 & 2.50 & 0 & 0 \\ \hline \hline
Single Observer & 0.38 & \cellcolor{yellow!35}0.53 & \cellcolor{yellow!35}1.65 & \cellcolor{yellow!35}0.80 & \cellcolor{yellow!35}0.64 & -8.49 & 6.19 & \cellcolor{yellow!35}3.48 \\ \hline
Center Prior & \cellcolor{yellow!35}0.45 & 0.38 & 0.92 & 0.78 & 0.51 & \cellcolor{yellow!35}0 & \cellcolor{yellow!35}1.24 & 3.72 \\ \hline
Permutation Control & 0.34 & 0.20 & 0.49 & 0.68 & 0.50 & -6.90 & 6.12 & 4.59 \\ \hline
Chance  & 0.33  & 0.00  & 0.00  & 0.50  & 0.50 & -1.24  & 2.09  & 6.35  \\ \hline
\end{tabular}
\caption{{\small Performance of saliency baselines (as pictured in Fig.~\ref{fig:vis}) with scores averaged over MIT300 benchmark images.}}
\label{tab:centerperf}
\vspace{-2em}
\end{table*}

\subsubsection{Earth Mover's Distance (EMD): \\Incorporating spatial distance into evaluation}
\label{sec:EMD}

\begin{figure*}
 \includegraphics[width=1\linewidth]{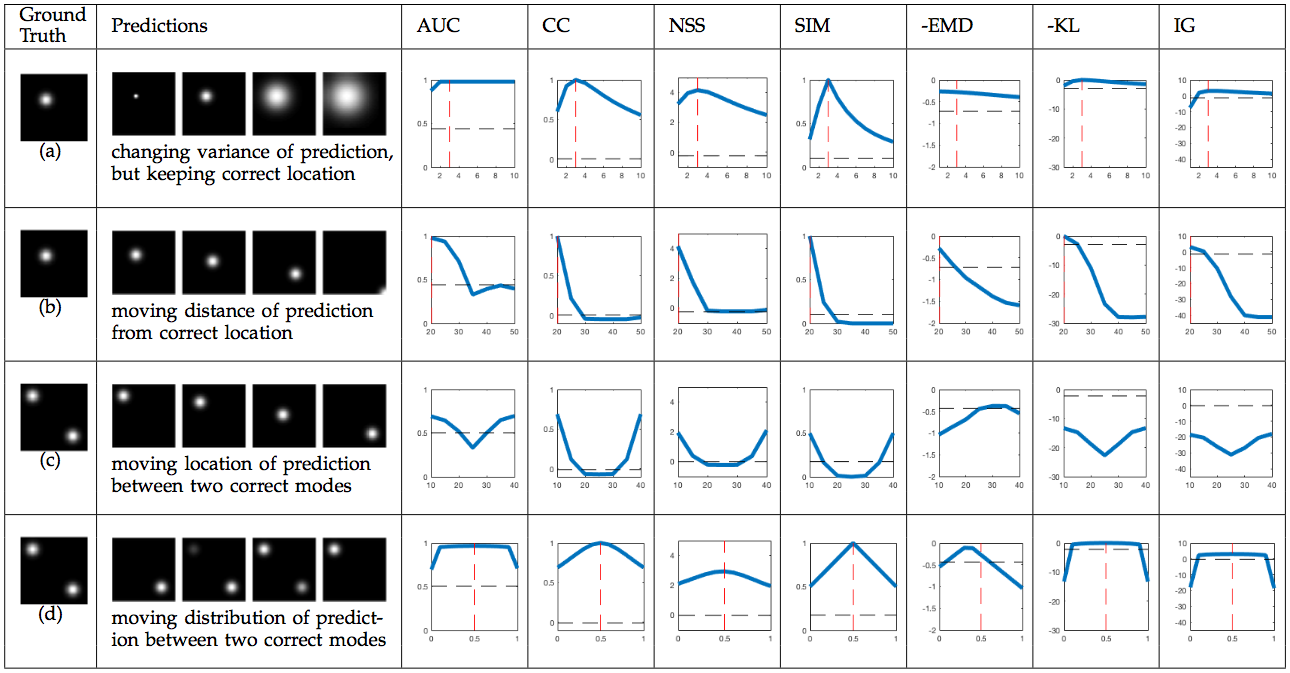}
    \caption{{\small We systematically varied parameters of a saliency map in order to quantify effects on metric scores. Each row corresponds to varying a single parameter value of the prediction: (a) variance, (b-c) location, and (d) relative weight. The x-axis of each subplot spans the parameter range, with the dotted red line corresponding to the ground truth parameter setting (if applicable). The y-axis is different across metrics but constant for a given metric. The dotted black line is chance performance. EMD and KL y-axes have been flipped so a higher y-value indicates better performance across all subplots.
}}
  \label{fig:synthetic1}
  \vspace{-1em}
\end{figure*}

All the metrics discussed so far have no notion of how spatially far away the prediction is from the ground truth. Accordingly, any map that has no pixel overlap with the ground truth will receive the same score of zero\footnote{Unless the model is properly regularized to compensate for uncertainty.}, regardless of how predictions are distributed (Fig.~\ref{fig:synthetic1}b). Incorporating a measure of spatial distance can broaden comparisons, and allow for graceful degradation when the ground truth measurements have position error.

The Earth Mover's Distance, \textbf{EMD}, measures the spatial distance between two probability distributions over a region. It was introduced as a spatially robust metric for image matching \cite{Rubner2000,Pele-eccv2008}.
Computationally, it is the minimum cost of morphing one distribution into the other. This is visualized in Fig.~\ref{fig:S_vs_EMD}d where in green are all the saliency map locations from which density needs to be moved, and in red are all the fixation map locations where density needs to be moved to. The total cost is the amount of density moved times the distance moved, and corresponds to brightness of the pixels in the visualization. It can be formulated as a transportation problem \cite{dantzig}.
We used the following linear time variant of EMD \cite{Pele-eccv2008}: 

\small
\begin{equation} \label{eq:EMD}
\begin{split}
\widehat{EMD}(P, Q^{D}) = \min_{\{f_{i j}\}} \sum_{i ,j}f_{i j}d_{i j} + | \sum_i P_i - \sum_j Q^{D}_j | \max_{i,j}d_{ij} \\
\mbox{    } \text{under the constraints:} \\
(1) f_{i j} \geq 0 \quad (2) \sum_j f_{i j} \leq P_i \quad (3) \sum_i f_{i j} \leq Q^{D}_j, \\
(4) \sum_{i, j} f_{i j} = \min(\sum_i P_i, \sum_j Q^{D}_j)  
\end{split}
\end{equation}
\normalsize

\noindent where each $f_{i j}$ represents the amount of density transported (or the \emph{flow}) from the $i$th supply to the $j$th demand and  $d_{i j}$ is the \emph{ground distance} between bin $i$ and bin $j$ in the distribution. Equation~\ref{eq:EMD} is therefore attempting to minimize the total amount of density movement such that the total density is preserved after the movement. Constraint (1) allows transporting density from $P$ to $Q^{D}$ and not vice versa. Constraint (2) prevents more density to be moved from a location $P_i$ than is there. Constraint (3) prevents more density to be deposited to a location $Q^{D}_j$ than is there. Constraint (4) is for feasibility: such that the amount of density moved does not exceed the total density found in either $P$ or $Q^{D}$. Solving this problem requires global optimization on the whole map, making this metric quite computationally intensive. 

A larger EMD indicates a larger difference between two distributions while an EMD of zero indicates that two distributions are the same. 
Generally, saliency maps that spread density over a larger area have larger EMD values (i.e., worse scores) as all the extra density has to be moved to match the ground truth map (Fig.~\ref{fig:S_vs_EMD}). EMD penalizes false positives proportionally to the spatial distance they are from the ground truth (Sec.~\ref{sec:misses_falsealarms}).

\newcommand{\cursw}{1.4cm}
\newcommand{\curs}{1.5cm}
\newcommand{\curss}{4cm}

\section{Analysis of metric behavior}
\label{sec:discussion}

This section contains a set of experiments to study the behavior of 8 different evaluation metrics, where we systematically varied properties of the input predictions to quantify the differential effects on metric scores. 
We focus on the metrics themselves, without assuming any optimization or regularization on the part of the inputs. This most closely reflects how evaluation is carried out on the MIT Saliency Benchmark, which does not place any restrictions on the format of the submitted saliency maps. As a result, our conclusions about the metrics should be informative for other applications, beyond saliency evaluation.

\newcommand{\h}{0.4cm}
\begin{table*}[tbh!]
\centering
\begin{tabular}{| l | c | c | c | c | c | c | c | c |}
\hline
 &AUC & sAUC & SIM & CC &KL & IG & NSS & EMD\\ \hline
\multicolumn{1}{| l |}{\textbf{Implementation}}&\multicolumn{8}{c |}{}\\ \hline
Bounded & \checkmark & \checkmark & \checkmark & \checkmark & &  & &  \\ \hline
Location-based, parameter-free & \checkmark & \checkmark & &  & & \checkmark  & \checkmark  & \\ \hline
Local computations, differentiable & & & \checkmark & \checkmark & \checkmark & \checkmark & \checkmark & \\ \hline
Symmetric & & & \checkmark & \checkmark & & &  & \checkmark\\ \hline
\multicolumn{1}{| l |}{\textbf{Behavior}}&\multicolumn{7}{c |}{} &\\ \hline
Invariant to monotonic transformations & \checkmark & \checkmark &  &  &  &  & &  \\ \hline
Invariant to linear transformations (contrast) & \checkmark & \checkmark &  & \checkmark & &  & \checkmark &  \\ \hline
Requires special treatment of center bias & & \checkmark & & & & \checkmark & & \\ \hline
Most affected by false negatives & & & \checkmark & & \checkmark & \checkmark & &\\ \hline
Scales with spatial distance &  &  &  & & & &  & \checkmark   \\ \hline
\end{tabular}
\caption{{\small Properties of the 8 evaluation metrics (with our specific implementations) considered in this paper.}}
\label{tab:properties}
\vspace{-1em}
\end{table*}


\subsection{Scoring baseline models}
\label{sec:baselines}

Comparing metrics on a set of baselines can be illustrative of metric behavior and be used to uncover the properties of saliency maps that drive this behavior. In Table~\ref{tab:centerperf} we include the scores of 4 baseline models and an upper bound for each metric. The \textbf{center prior} model is a symmetric Gaussian stretched to the aspect ratio of the image, so each pixel's saliency value is a function of its distance from the center (higher saliency closer to center). Our \textbf{chance} model assigns a uniform value to each pixel in the image. An alternative chance model that also factors in the properties of a particular dataset is called a \textbf{permutation control}: it is computed by randomly selecting a fixation map from another image. It has the same image-independent properties as the ground truth fixation map for the image since it has been computed with the same blur and scale. 
The \textbf{single observer} model uses the fixation map from one observer to predict the fixations of the remaining observers (1~predicting $n-1$). We repeated this leave-one-out procedure and averaged the results across all observers. 

To compute an upper bound for each metric we measured how well the fixations of $n$ observers predict the fixations of another group of $n$ observers, varying $n$ from 1 to 19 (half of the total 39 observers). Then we fit these prediction scores to a power function to obtain the limiting score of \textbf{infinite observers}. The details of this computation can be found in the appendix. This is useful to obtain dataset-specific bounds for metrics that are not otherwise bounded (i.e. NSS, EMD, KL, IG), and to provide realistic bounds that factor in dataset-specific human consistency for metrics where the theoretical bound may not be reachable (i.e. AUC, sAUC). 

There is a divergent behavior in the way the metrics score a center prior model relative to a single observer model. The center prior captures dataset-specific, image-independent properties; while the single observer model captures image-specific properties but might be missing properties that emulate average viewing behavior. In particular, the single observer model is quite sparse and so achieves worse scores according to the KL, IG, and SIM metrics. 

Similarly, we compare the chance and permutation control models. Both are image-independent. However, the chance model is also dataset-independent, while the permutation control model captures some dataset-specific properties. The CC, NSS, AUC, and EMD scores are significantly higher for the permutation control, pointing to the importance under these metrics, of capturing the properties of a particular dataset (including center bias, blur, and scale). On the other hand, KL and IG are sensitive to insufficient regularization. As a result, the permutation control model, which has more zero values, fares worse than the chance model.

One possible meta-measure for selecting metrics for evaluation is how much better one baseline is over another (e.g., \cite{emami2013selection,margolin2014evaluate,pont2016supervised}). However, the optimal ranking of baselines is likely to be different across applications: in some cases, it may be useful to accurately capture systematic viewing behaviors if nothing else is known, while in another setting, specific points of interest are more relevant than viewing behaviors.



\subsection{Treatment of false positives and negatives}
\label{sec:misses_falsealarms}

Different metrics place different weights on the presence of false positives and negatives in the predicted saliency relative to the ground truth. 
To directly compare the extent to which metrics penalize false negatives, we performed a series of systematic tests. Starting with the ground truth fixation map, we progressively removed different amounts of salient pixels: pixels with a saliency value above the mean map value were selected uniformly at random and set to 0. We then evaluated the similarity of the resulting map to the original ground truth map and measured the drop in score with 25\%, 50\%, and 75\% false negatives. 
To make comparison across metrics possible, we normalized this change in score by the score difference between the infinite observer limit and chance. We call this the \textbf{chance-normalized score}. For instance, for the AUC-Judd metric the upper limit is 0.92, chance is at 0.50, and the score with 75\% false negatives is 0.67. The chance-normalized score is: $100\% \times (0.92-0.67)/(0.92-0.50) = 60\%$. Values for the other metrics are available in Table~\ref{tab:misses_perf}. \\

\newcommand{\hn}{0.58cm}
\newcommand{\hnn}{0.68cm}
\begin{table}
\centering
\begin{tabular}{| m{0.62cm} | m{\hn} | m{\hn} | m{\hn} | m{\hn} | m{\hn} | m{0.73cm} | m{\hnn} |}
\hline
Map & EMD & CC & NSS & AUC & SIM  & IG & KL    \\ 
 &  $\downarrow$ &  $\uparrow$ &  $\uparrow$ &  $\uparrow$ &  $\uparrow$ &  $\uparrow$ &  $\downarrow$   \\ \hline
Orig & 0.00 & 1.00 & 3.29 & 0.92 & 1.00 & 2.50 & 0.00 \\ 
\rowcolor{lightgray}
 &  (0\%) &  (0\%) &  (0\%) &  (0\%) &  (0\%) &  (0\%) & (0\%) \\ \hline
-25\% & 0.13 & 0.85 & 2.66 & 0.85 & 0.78 & -1.78 & 2.55 \\ 
\rowcolor{lightgray}
 &  (2\%) &  (15\%) &  (19\%) &  (17\%) &  (33\%) & (114\%) &  (122\%)  \\ \hline
-50\% & 0.16 & 0.70 & 2.18 & 0.77 & 0.59 & -6.35 & 5.64 \\ 
\rowcolor{lightgray}
 &  (3\%) &  (30\%) &  (34\%) &  (36\%) &  (61\%) & (237\%) &  (270\%)  \\ \hline
-75\% & 1.09  & 0.50 & 1.57 & 0.67 & 0.45  & -10.65 & 8.18 \\ 
\rowcolor{lightgray}
&  (17\%) &  (50\%) &  (52\%) &  (60\%) &  (82\%)  & (352\%) &  (391\%) \\ \hline
\end{tabular}
\caption{{\small Metrics have different sensitivities to false negatives. We sorted these metrics in order of increasing sensitivity to 25\%, 50\%, and 75\% false negatives, where EMD is least, and KL is most, sensitive. Scores are averaged over all MIT300 fixation maps. Below each score is the percentage drop in performance from the metric's limit, normalized by the percentage drop to chance level. 
}}
\label{tab:misses_perf}
\vspace{-1em}
\end{table}

\noindent\textbf{KL, IG, and SIM are most sensitive to false negatives:}
If the prediction is close to zero where the ground truth has a non-zero value, the penalties can grow arbitrarily large under KL, IG, and SIM. These metrics penalize models with false negatives significantly more than false positives. In Table~\ref{tab:misses_perf}, KL and IG scores drop below chance levels with only 25\% false negatives. Another way to look at this is that these metrics' sensitivity to regularization drives their evaluations of models. KL and IG scores will be low for sparse and poorly regularized models. \\



\noindent\textbf{AUC ignores low-valued false positives:}
AUC scores are a function of which level sets the false positives fall into - where false positives in the first few level sets are penalized most, but false positives in the last level set do not have a large impact on performance. Models with many low-valued false positives (e.g., Fig.~\ref{fig:AUC_vs_NSS}) do not incur large penalties. Saliency maps that place different amounts of density but at the correct (fixated) locations will receive similar AUC scores (Fig.~\ref{fig:synthetic1}d). 
\\ 

\noindent\textbf{NSS and CC are equally affected by false positives and negatives:}
During the normalization step of NSS, a few false positives will be washed out by the other saliency values and will not significantly affect the saliency values at fixated locations. However, as the number of false positives increases, they begin to have a larger influence on the normalization calculation, driving the overall NSS score down.

By construction, CC has a symmetric treatment of false positives and negatives. However, NSS is highly related to CC, and can be viewed as a discrete approximation (see appendix). NSS behavior will be very similar to CC, including the treatment of false positives and negatives.\\

\noindent\textbf{EMD's penalty depends on spatial distance:}
EMD is least sensitive to uniformly-occurring false negatives (e.g., Table~\ref{tab:misses_perf}) because the EMD calculation can redistribute saliency values from nearby pixels to compensate. 
However, false negatives that are spatially far away from any predicted density are highly penalized. Similarly, EMD's penalty for false positives depends on their spatial location relative to the ground truth, in that false positives close to ground truth locations can be redistributed to those locations at low cost, but distant false positives are highly penalized (Fig.~\ref{fig:S_vs_EMD}). 

\subsection{Systematic viewing biases}
\label{sec:systematicviewing}

Common to many images is a higher density of fixations in the center of the image compared to the periphery, a function of both photographer bias (i.e., centering the main subject) and observer viewing biases. The effect of center bias on model evaluation has received much attention  \cite{BylinskiiOpinion,Einhauser2008JOV,le2006coherent,Parkhurst2002107,Renninger05aninformation,Tatler2007JOV,Tatler2005,Zhao10032011}. In this section we discuss center bias in the context of the metrics in this paper. \\


\noindent\textbf{sAUC penalizes models that include center bias:}
The sAUC metric samples negatives from other images, which in the limit of many images corresponds to sampling negatives from a central Gaussian. For an image with a strong central viewing bias, both positives and negatives would be sampled from the same image region, and a correct prediction would be at chance (Fig.~\ref{fig:centerbias}).
The sAUC metric prefers models that do not explicitly incorporate center bias into their predictions. For a fair evaluation under sAUC, models need to operate under the same assumptions, or else their scores will be dominated by whether or not they incorporate center bias. 
\\

\noindent\textbf{IG provides a direct comparison to center bias:}
Information gain over a center prior baseline provides a more intuitive way to interpret model performance relative to center bias. If a model can not explain fixation patterns on an image beyond systematic viewing biases, such a model will have no gain over a center prior.\\

\noindent\textbf{EMD spatially hedges its bets:}
The EMD metric prefers models that hedge their bets if all the ground truth locations can not be accurately predicted (Fig.~\ref{fig:synthetic1}c). For instance, if an image is fixated in multiple locations, EMD will favor a prediction that falls spatially between the fixated locations instead of one that captures a subset of the fixated locations (contrary to the behavior of the other metrics). \\

A center prior is a good approximation of average viewing behavior on images under laboratory conditions, where an image is projected for a few seconds on a computer screen in front of an observer \cite{Bindermann2010}.  
A dataset-specific center prior emerges when averaging fixations over a large set of images.
Knowing nothing else about image content, the center bias can act as a simple model prior. 
Overall if the goal is to predict natural viewing behavior on an image, center bias is part of the viewing behavior and discounting it entirely may be suboptimal. 
However, different metrics make different assumptions about the models: sAUC penalizes models that include center bias, while IG expects center bias to already be optimized. These differences in metric behaviors have lead to differences in whether models include or exclude center bias (e.g.~\cite{Judd_2012,mit-saliency-benchmark}). As a result, model rankings according to a particular metric can often be dominated by the differences in modeled center bias (Sec.~\ref{sec:corrs}). 


\subsection{Relationship between metrics}
\label{sec:corrs}

\begin{figure}
\centering
\includegraphics[width=0.8\linewidth]{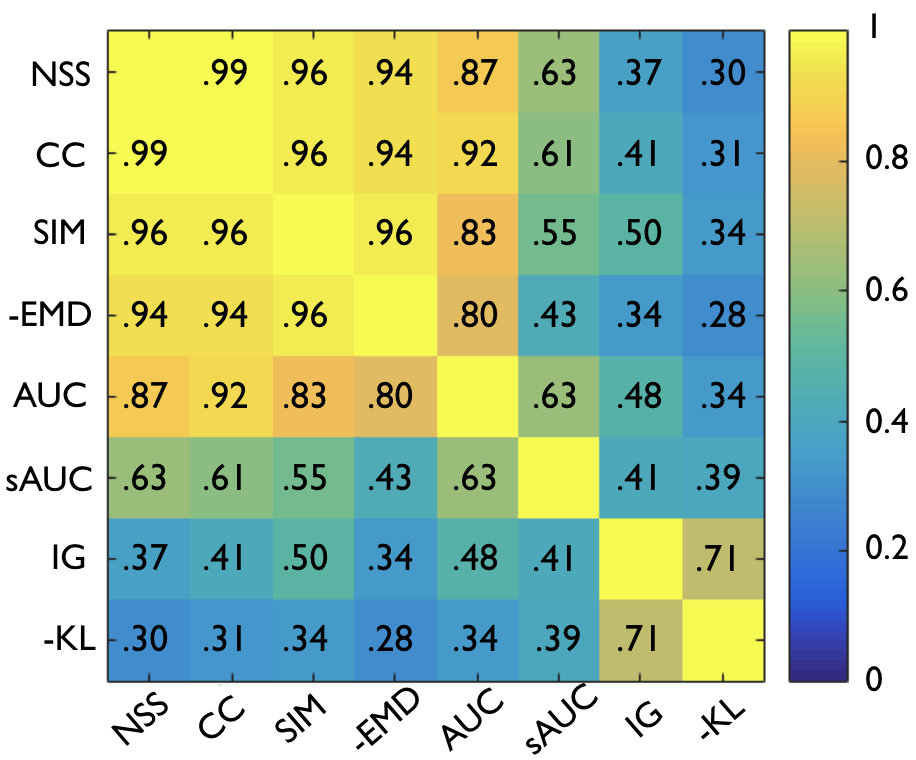}
\caption{{\small We sort the saliency models on the MIT300 benchmark individually by each metric, and then compute the Spearman rank correlation between the model orderings of every pair of metrics. The first 5 metrics listed are highly correlated with each other. KL and IG are highly correlated with each other and most uncorrelated with the other metrics, due to their high sensitivity to zero-valued predictions at fixated locations. The sAUC metric is also different from the others because it specifically penalizes models that have a center bias.}}
\label{fig:corr}
\vspace{-1em}
\end{figure}

As saliency metrics are often used to rank saliency models, we can measure how correlated the rankings are across metrics. This analysis will indicate whether metrics favor or penalize similar behaviors in models.
We sort model performances according to each metric and compute the Spearman rank correlation between the model orderings of every pair of metrics to obtain the correlation matrix in Fig.~\ref{fig:corr}.
The pairwise correlations between NSS, CC, AUC, EMD, and SIM range from 0.76 to 0.98. Because of these high correlations, we call this the \textbf{similarity cluster} of metrics. CC and NSS are most highly correlated due to their analogous computations, as are KL and IG (see appendix). 

Driven by extreme sensitivity to false negatives, KL, IG, and SIM rank saliency models differently than the similarity cluster. 
Viewed another way, these metrics are worse behaved if saliency models are not properly regularized. For these metrics, a zero-valued prediction is interpreted as an impossibility of fixations at that location, while for the other metrics, a zero-valued prediction is treated as less salient. 
These metrics have a natural probabilistic interpretation and are appropriate in cases where missing any ground truth fixation locations should be highly penalized, such as for detection applications (Sec.~\ref{sec:applications}). Changing the regularization constant $\epsilon$ in the metric computations (Eq.~\ref{eq:IG},\ref{eq:KL}) or regularizing the saliency models prior to evaluation (as in \cite{kummerer2015information}) can reduce score differences between KL, IG, and the similarity cluster.

Although EMD is the only metric that takes into account spatial distance, it nevertheless ranks saliency models similarly as the other similarity cluster metrics. This is likely the case for two reasons: (i) like the similarity cluster metrics, EMD is also center biased (Table~\ref{tab:centerperf}, Sec.~\ref{sec:systematicviewing}) and (ii) current model mistakes are often a cause of completely incorrect prediction rather than imprecise localization (note: as models continue to improve, this might change).


Shuffled AUC (sAUC) has low correlations with other metrics because it modifies how predictions at different spatial locations on the image are treated. 
A model with more central predictions will be ranked lower than a model with more peripheral predictions (Fig.~\ref{fig:centerbias}). Shuffled AUC assumes center bias has not been modeled, and penalizes models where it has.
 For these reasons, sAUC has been disfavored by some evaluations \cite{bruce2015computational,li2015data,LeMeur2013}.
An alternative is optimizing models to include a center bias \cite{Judd_2012,Judd_2009,kummerer2014close,kummerer2015information,Parkhurst2003,Zhao10032011}. In this case, the metric can be ambivalent to any model or dataset biases. 

Saliency metrics are much more correlated once models are optimized for center bias, blur, and scale \cite{Judd_2012,kummerer2014close,kummerer2015information}. As a result, the differences between the metrics in Fig.~\ref{fig:corr} are largely driven by how sensitive the metrics are to these model properties. It is therefore valuable to know if different models make similar modeling assumptions in order to interpret saliency rankings meaningfully across metrics.

\subsection{Comparisons to related work:}

Riche et al.~\cite{RicheICCV2013} correlated metric scores on another saliency dataset and found that KL and sAUC are most different from the other metrics, including AUC, CC, NSS, and SIM, which formed a single cluster. 
We can explain this finding, since KL and sAUC make stronger assumptions about saliency models: KL assumes saliency models have sufficient regularization (otherwise false negatives are severely penalized) and sAUC assumes the model does not have a built-in center bias.
Both Riche et al.~\cite{RicheICCV2013} and our results show that these assumptions do not always hold for the commonly evaluated saliency models, leading to divergent rankings across metrics.


Emami and Hoberock~\cite{emami2013selection} used human consistency to compare 9 metrics. In discriminating between human saliency maps and random saliency maps, they found that NSS and CC were the best, and KL the worst. 
This is similar to the analysis in Sec.~\ref{sec:baselines}.

Li et al.~\cite{li2015data} used crowd-sourced experiments to measure which metric best corresponds to human perception. 
The authors noted that human perception was driven by the most salient locations, the compactness of salient locations (i.e., low false positives), and a similar number of salient regions as the ground truth. 
As a result, the perception-based ranking most closely matched that of NSS, CC, and SIM, and was furthest from KL and EMD. 
However, the properties that drive human perception could be different than the properties desired for other applications of saliency. For instance, for evaluating probabilistic saliency maps, proper regularization and the scale of the saliency values (including very small values) can significantly affect evaluation. For these cases, perception-based metrics might not be as appropriate.

We propose that the assumptions underlying different models and metrics be considered more carefully, and that the different metric behaviors and properties enter into the decision of which metrics to use for evaluation (Table~\ref{tab:properties}). 

\section{Recommendations for designing a saliency benchmark}
\label{sec:designSalBenchmark}

Saliency models have evolved significantly since the seminal IttiKoch model~\cite{Koch1985,Walther2006} and the original notions of saliency. Evaluation procedures, saliency datasets, and benchmarks have adapted accordingly. Given how many different metrics and models have emerged, it is becoming increasingly necessary to systematize definitions and evaluation procedures to make sense of the vast amount of new data and results~\cite{BylinskiiOpinion}. The MIT Saliency Benchmark is a product of this evolution of saliency modeling; an attempt to capture the latest developments in models and metrics. However, as saliency continues to develop as a research area, larger more specialized datasets may become more appropriate. 
Based on our experience with the MIT Saliency Benchmark, we provide some recommendations for future saliency benchmarks.

\subsection{Defining expected input}

As observed in the previous section, some of the inconsistencies in how metrics rank models are due to differing assumptions that saliency models make. This problem has been emphasized by K{\"u}mmerer et al. \cite{kummerer2014close,kummerer2015information}, who argued that if models were explicitly designed and submitted as \emph{probabilistic models}, then some ambiguities in evaluation would disappear. For instance, a probability of zero in a probabilistic saliency map assumes that a fixation in a region is impossible; under alternative definitions, a value of zero might only mean that a fixation in a particular region is less likely. Metrics like KL, IG, and SIM are particularly sensitive to zero values, so models evaluated under these metrics would benefit from being regularized and optimized for scale. Similarly, knowing whether evaluation will be performed with a metric like sAUC should affect whether center bias is modeled, because this design decision would be penalized under this metric. 
A saliency benchmark should specify what definition of saliency is assumed, what kind of saliency map input is expected, and how models will be evaluated. The appendix includes additional considerations. 

\subsection{Handling dataset bias}

In saliency datasets, dataset bias occurs when there are systematic properties in the ground-truth data that are dataset-specific but image-independent. 
Most eye-tracking datasets have been shown to be center biased, containing a larger number of fixations near the image center, across different image types, videos, and even observer tasks~\cite{borji2013quantitative,BorjiICCV2013,Canosa2003,Clarke2014,Itti_Baldi06nips,JuddJOV2011}. 
Center bias is a function of multiple factors, including photographer bias and observer bias, due to the viewing of fixed images in a laboratory setting~\cite{Tatler2007JOV,wloka2013overt}.
As a result, some models have a built-in center bias (e.g.,~Judd~\cite{Judd_2009}), some metrics penalize center bias (e.g.,~sAUC), and some benchmarks optimize models with center bias prior to evaluation (e.g.,~LSUN~\cite{LSUN}). These different approaches result from a disagreement in where systematic biases should be handled: at the level of the dataset, model, or evaluation. For transparency, saliency benchmarks should specify whether the submitted models are expected to incorporate center bias, or if dataset-specific center bias will be accounted for and subtracted during evaluation. In the former case, the benchmark can provide a training dataset on which to optimize center bias and other image-independent properties of the ground truth dataset (e.g., blur, scale, regularization), or else share these parameters directly.

The MIT Saliency Benchmark provides the MIT1003 dataset~\cite{Judd_2009} as a training set to optimize center bias and blur parameters, and for histogram matching (scale regularization)\footnote{Associated code is provided at \url{https://github.com/cvzoya/saliency/tree/master/code_forOptimization}.}. Both MIT300 and MIT1003 have been collected using the same eye tracker setup, so the ground truth fixation data should have similar distribution characteristics, and parameter choices should generalize across these datasets. 

The first saliency models were not designed with these considerations in mind, so when compared to models that had incorporated center bias and other properties into saliency predictions, the original models were at a disadvantage. However, the availability of saliency datasets has increased, and many benchmarks provide training data from which systematic parameters can be learned~\cite{mit-saliency-benchmark,LSUN,jiang2015salicon}. Many modern saliency models are a result of this data-driven approach. Over the last few years, we have seen fewer differences across saliency models in terms of scale, blur, and center bias~\cite{mit-saliency-benchmark}.


\subsection{Defining a task for evaluation}
\label{sec:applications}


Saliency models are often designed to predict general task-free saliency, assigning a value of saliency or importance to each image pixel, largely independent of the end application. 
Saliency is often motivated as a useful representation for image processing applications such as image re-targeting, compression, and transmission, object and motion detection, and image retrieval and matching \cite{borji2013state,Judd_thesis}. However, if the end goal is one of these applications, then it might be easier to directly train a saliency model for the relevant task, rather than for task-free fixation prediction. 
Task-based, or application-specific, saliency prediction is not yet very common. Relevant datasets and benchmarks are yet to be designed.
Evaluating saliency models on specific applications requires choosing metrics that are appropriate to the underlying task assumptions and expected input. 

Consider a detection application of saliency such as object and motion detection, surveillance, localization and mapping, and segmentation \cite{Achanta2008Book,chang2010mobile,frintrop2010general,frintrop2007simultaneous,klein2011center,Liu2007CVPR,navalpakkam2006integrated,yun2013studying}. For such an application, a saliency model may be expected to produce a probability density of possible object locations, and be highly penalized if a target is missed. For this kind of probabilistic target detection, AUC, KL, and IG would be appropriate. EMD might be useful if some location invariance is permitted.

Applications including adaptive image and video compression and progressive transmission \cite{Geisler1998,itti2004automatic,ma2005generic,Wang03foveationscalable}, thumbnailing \cite{marchesotti2009framework,suh2003automatic}, content-aware image re-targeting and cropping \cite{achanta2009saliency,Avidan2007ACM,Rub2008Siggraph,Santella2006CHI,wang2011saliency}, rendering and visualization \cite{kim2006saliency,longhurst2006gpu}, collage \cite{goferman2010puzzle,wang2006picture} and artistic rendering \cite{decarlo2002stylization,Judd_2009} require ranking (by importance or saliency) different image regions. For these applications, when it is valuable to know how much more salient a given image region is than another, an evaluation metric like AUC (that is ambivalent to monotonic transformations of the input map) is not appropriate. Instead, NSS or SIM would provide a more useful evaluation.

\subsection{Selecting metrics for evaluation} \label{sec:selectingMetrics}

A goal of this paper has been to show how metrics behave under different conditions. This can help guide the selection of metrics for saliency benchmarks, depending on the assumptions that are made (e.g., whether the models are probabilistic, whether center bias is accounted for, etc.). A saliency benchmark should specify any assumptions that can be made along with the expected saliency map format. 

The MIT Saliency Benchmark assumes that all fixation behavior is part of the saliency modeling: including any systematic dataset parameters (e.g., blur, scale, etc.). Capturing viewing biases is part of the modeling requirements. Metrics like shuffled AUC will penalize models that have a strong center bias. Saliency models submitted are not necessarily probabilistic, so they might be unfairly evaluated by the KL, IG, and SIM metrics that penalize zero values (false negatives), unless they are first regularized and pre-processed as in K{\"u}mmerer et al. \cite{kummerer2015information}. 
AUC has begun to saturate on the MIT Saliency Benchmark and is becoming less capable of discriminating between different saliency models \cite{bylinskii2016should}. This is because AUC is ambivalent to monotonic transformations. However, for certain saliency applications it might be valuable to know exactly how much more salient a given image region is than another, and not just their relative saliency ranks. 
Of the remaining metrics, the Earth Mover's Distance (EMD) is computationally expensive to compute and difficult to optimize for.
Given all of this, for a benchmark operating under the same assumptions as the MIT Saliency Benchmark, we recommend reporting either CC or NSS. Both make limited assumptions about input format, and treat false positives and negatives symmetrically.
For a benchmark intended to evaluate saliency maps as probability distributions, IG and KL would be good choices; IG specifically measures prediction performance beyond systematic dataset biases.

\begin{table*}
\centering
\begin{tabular}{| p{3cm} | p{13cm} |}
\hline
Metric & Quick take-aways \\ \hline
Area under ROC Curve (AUC) & Historically the most commonly-used metric for saliency evaluation. Invariant to monotonic transformations. Driven by high-valued predictions and largely ambivalent of low-valued false positives. Currently saturating on standard saliency benchmarks~\cite{mit-saliency-benchmark,bylinskii2016should}. Good for detection applications. \\ \hline
Shuffled AUC (sAUC) & A version of AUC that compensates for dataset bias by scoring a center prior at chance. Most appropriate in evaluation settings where the saliency model is not expected to account for center bias. Otherwise, has similar properties to AUC.\\ \hline
Similarity (SIM) & An easy and fast similarity computation between histograms. Assumes the inputs are valid distributions. More sensitive to false negatives than false positives. \\ \hline
Pearson's Correlation Coefficient (CC) &  A linear correlation between the prediction and ground truth distributions. Treats false positives and false negatives symmetrically. \\ \hline
Normalized Scanpath Saliency (NSS) & A discrete approximation of CC that is additionally parameter-free (operates on raw fixation locations). Recommended for saliency evaluation. \\ \hline
Earth Mover's Distance (EMD) & The only metric considered that scales with spatial distance. Can provide a finer-grained comparison between saliency maps. Most computationally intensive, non-local, hard to optimize. \\ \hline
Kullback-Leibler divergence (KL) & Has a natural interpretation where goal is to approximate a target distribution. Assumes input is a valid probability distribution with sufficient regularization. Mis-detections are highly penalized.\\ \hline
Information Gain (IG) & A new metric introduced by~\cite{kummerer2014close,kummerer2015information}. Assumes input is a valid probability distribution with sufficient regularization. Measures the ability of a model to make predictions above a baseline model of center bias. Otherwise, has similar properties to KL.\\ \hline
\end{tabular}
\caption{{\small A brief overview of the metric analyses and discussions provided in this paper, highlighting some of the key properties, features, and applications of different evaluation metrics.}}
\label{tab:takeaways}
\vspace{-1em}
\end{table*}

\section{Conclusion}

We provided an analysis of the behavior of 8 evaluation metrics to make sense of the differences in saliency model rankings according to different metrics.
Properties of the inputs affect metrics differently: how the ground truth is represented; whether the prediction includes dataset bias; whether the inputs are probabilistic; whether spatial deviations exist between the prediction and ground truth. 
Knowing how these properties affect metrics, and which properties are most important for a given application can help with metric selection for saliency model evaluation.
Other considerations for metric selection include whether the metric computations are expensive, local, and differentiable, which would influence whether a metric is appropriate for model optimization. Take-aways about the metrics are included in Table~\ref{tab:takeaways}.

We considered saliency metrics from the perspective of the MIT Saliency Benchmark, which does not assume that saliency models are probabilistic as in~\cite{kummerer2014close,kummerer2015information}, but does assume that all systematic dataset biases (including center bias, blur, scale) are taken care of by the model. Under these assumptions we found that the Normalized Scanpath Saliency (NSS) and Pearson's Correlation Coefficient (CC) metrics provide the fairest comparison. Being closely related mathematically, their rankings of saliency models are highly correlated, and reporting performance using one of them is sufficient. However, under alternative assumptions and definitions of saliency, another choice of metrics may be more appropriate. Specifically, if saliency models are evaluated as probabilistic models, then KL-divergence and Information Gain (IG) are recommended. Arguments for why it might be preferable to define and evaluate saliency models probabilistically can be found in \cite{kummerer2014close,kummerer2015information}. Specific tasks and applications may also call for a different choice of metrics. For instance, AUC, KL, and IG are appropriate for detection applications, as they penalize target detection failures. However, where it is important to evaluate the relative importance of different image regions, such as for image-retargeting, compression, and progressive transmission, metrics like NSS or SIM are a better fit.

In this paper we discussed the influence of different assumptions on the choice of appropriate metrics. We provided recommendations for new saliency benchmarks, such that if designed with explicit assumptions from the start, evaluation can be more transparent and reduce confusion in saliency evaluation. 
We also provide code for evaluating and visualizing the metric computations\footnote{\url{http://saliency.mit.edu/downloads.html}} to add further transparency to model evaluation and to allow researchers a finer-grained look into metric computations, to debug saliency models and visualize the aspects of saliency models driving or hurting performance.

 \section*{Acknowledgments}
  The authors would like to thank Matthias K{\"u}mmerer and other attendees of the saliency tutorial at ECCV 2016 for helpful discussions about saliency evaluation\footnote{\url{http://saliency.mit.edu/ECCVTutorial/ECCV_saliency.htm}}.
  Thank you also to the anonymous reviewers for many detailed suggestions. 
  ZB was supported by a postgraduate scholarship (PGS-D) from the Natural Sciences and Engineering Research Council of Canada. 
  Support to AO, AT, and FD was provided by the Toyota Research Institute / MIT CSAIL Joint Research Center.

\bibliography{main_red}
\bibliographystyle{plain}

%
\begin{IEEEbiography}[{\includegraphics[width=1in,height=1.25in,clip,keepaspectratio]{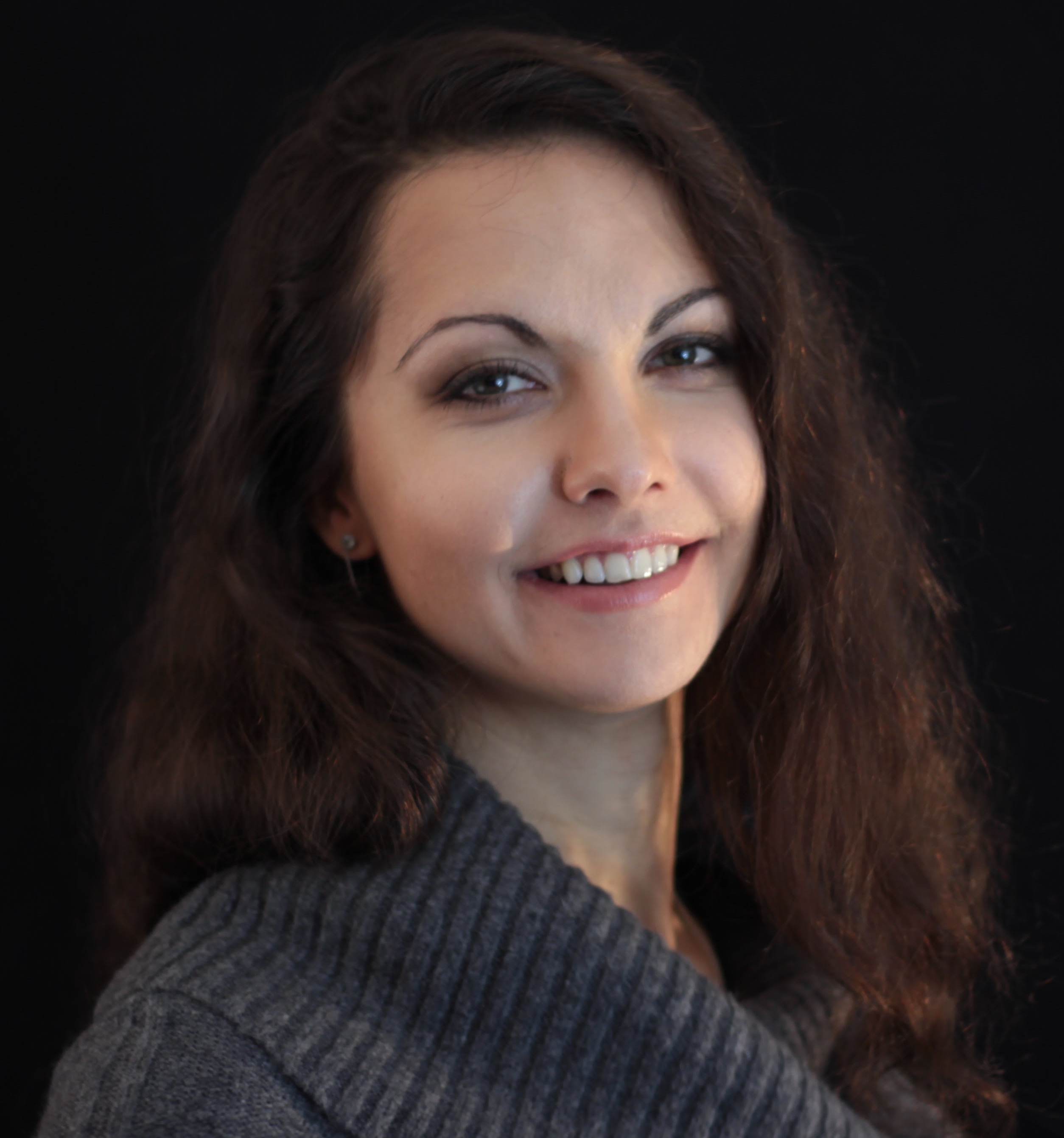}}]{Zoya Bylinskii} is a Ph.D. candidate in Computer Science at the Massachusetts Institute of Technology, advised by Fr\'edo Durand and Aude Oliva. She received a B.S. degree in Computer Science and Statistics at the University of Toronto, followed by an M.S. degree in Computer Science from MIT under the supervision of Antonio Torralba and Aude Oliva. Zoya works on topics at the interface of human and computer vision, on computational perception and cognition. 
She is an Adobe Research Fellow (2016), a recipient of the Natural Sciences and Engineering Research Council of Canada's (NSERC) Postgraduate Doctoral award (2014-6), Julie Payette NSERC Scholarship (2013), and a finalist for Google's Anita Borg Memorial Scholarship (2011).
\end{IEEEbiography}

\begin{IEEEbiography}[{\includegraphics[width=1in,height=1.25in,clip,keepaspectratio]{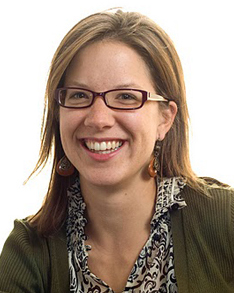}}]{Tilke Judd} is a Product Manager at Google in Zurich. She received a B.S. degree in Mathematics from Massachusetts Institute of Technology (MIT) followed by an M.S. degree in Computer Science and a Ph.D. in Computer Science from MIT in 2007 and 2011 respectively, supervised by Fr\'edo Durand and Antonio Torralba.  During the summers of 2007 and 2009 she was an intern with Google and Industrial Light and Magic respectively.  She was awarded a National Science Foundation Fellowship for 2005-2008 and a Xerox Graduate Fellowship in 2008.
\end{IEEEbiography}

\begin{IEEEbiography}[{\includegraphics[width=1in,height=1.25in,clip,keepaspectratio]{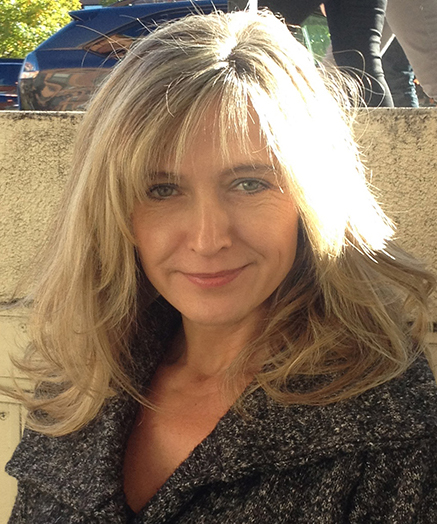}}]{Aude Oliva} is a Principal Research Scientist at the MIT Computer Science and Artificial Intelligence Laboratory (CSAIL). After a French baccalaureate in Physics and Mathematics and a B.Sc. in Psychology, Aude Oliva received two M.Sc. degrees and a Ph.D in Cognitive Science from the Institut National Polytechnique of Grenoble, France. She joined the MIT faculty in the Department of Brain and Cognitive Sciences in 2004 and CSAIL in 2012. Her research on vision and memory is cross-disciplinary, spanning human perception and cognition, computer vision, and human neuroscience. She is the recipient of a National Science Foundation CAREER Award in Computational Neuroscience (2006), the Guggenheim fellowship in Computer Science (2014), and the Vannevar Bush faculty fellowship in Cognitive Neuroscience (2016).
\end{IEEEbiography}

\begin{IEEEbiography}[{\includegraphics[width=1in,height=1.25in,clip,keepaspectratio]{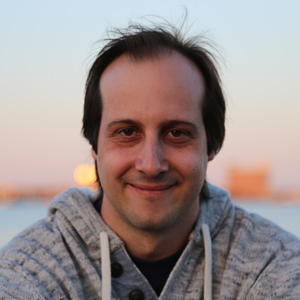}}]{Antonio Torralba} is a Professor of Electrical Engineering and Computer Science at the Massachusetts Institute of Technology. He received the degree in telecommunications engineering from Telecom BCN, Spain, in 1994 and the Ph.D. degree in signal, image, and speech processing from the Institut National Polytechnique de Grenoble, France, in 2000. From 2000 to 2005, he spent postdoctoral training at the Brain and Cognitive Science Department and the Computer Science and Artificial Intelligence Laboratory, MIT. He received the 2008 National Science Foundation (NSF) Career award, the best student paper award at the IEEE Conference on Computer Vision and Pattern Recognition (CVPR) in 2009, and the 2010 J. K. Aggarwal Prize from the International Association for Pattern Recognition (IAPR).
\end{IEEEbiography}

\begin{IEEEbiography}[{\includegraphics[width=1in,height=1.25in,clip,keepaspectratio]{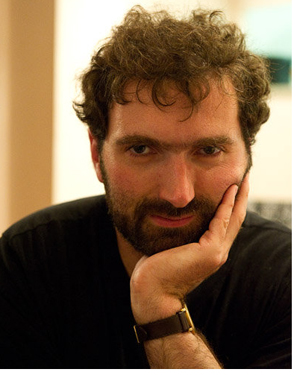}}]{Fr\'edo Durand} is a professor in Electrical Engineering and Computer Science at the Massachusetts Institute of Technology, and a member of the Computer Science and Artificial Intelligence Laboratory (CSAIL). He received his PhD from Grenoble University, France, in 1999, supervised by Claude Puech and George Drettakis. From 1999 till 2002, he was a post-doc in the MIT Computer Graphics Group with Julie Dorsey.
\end{IEEEbiography}



\setcounter{section}{0}
\renewcommand\thesection{\Alph{section}}

\section{Appendix}

\input{just_appendix}

\ifCLASSOPTIONcaptionsoff
  \newpage
\fi

\end{document}

%% file: just_appendix.tex
\subsection{Evaluation setup: data collection}

Images for the MIT300 dataset were obtained from Flickr Creative Commons and personal photo collections.
Eye movements were collected using a table-mounted, video-based ETL 400 ISCAN eye tracker which recorded observers' gaze paths at 240Hz. The average calibration error was less than one degree of visual angle. 
Each image was presented for 2 seconds at a maximum dimension of 1024 pixels and the second dimension between 457-1024 pixels (mode: 768 pixels). 
The task instruction was: \emph{"You will see a series of 300 images. Look closely at each image. After viewing the images you will have a memory test: you will be asked to identify whether or not you have seen particular images before"}. This was used to motivate participants to pay attention, but no memory test was used.
Images were separated by a 500 ms fixation cross. During pre-processing, the first fixation on each image was thrown out to reduce the center-biasing effects of the fixation cross. 
A list of alternative eyetracking datasets with different experimental setups, tasks, images, and exposure durations is available at \url{http://saliency.mit.edu/datasets.html}.

\subsection{Metric computation}

\noindent\textbf{Location-based versus distribution-based metrics:}

The particular \emph{implementations} of the metrics we use can be categorized as either location-based or distribution-based, as presented in the paper. However, there are implementations of AUC and NSS that require the ground truth to be a distribution \cite{LeMeur2013}. In these implementations, the ground truth distribution is then pre-processed into a binary map by thresholding at a fixed, often arbitrary value. This requires an additional parameter for the metric computation. Our parameter-free, location-based implementations of AUC and NSS are more commonly used for saliency evaluation. \\  \\



\noindent\textbf{Sampling thresholds for the ROC curve:}

The ROC curve is obtained by plotting the true positive rate against the false positive rate at various thresholds of the saliency map. 
Choosing how to sample thresholds to approximate the continuous ROC curve is an important implementation consideration.
A saliency map is first normalized so all saliency values lie between 0 and 1. In the AUC-Judd implementation, each distinct saliency map value is used as a threshold,
so this sampling strategy provides the most accurate approximation to the continuous curve. 
To ensure that enough threshold samples are taken, the saliency map is first jittered by adding a tiny random value to each pixel, thus preventing large uniform regions of one value in the saliency map.

In the AUC-Borji implementation the threshold is sampled at a fixed step size (from 0 to 1 by increments of 0.1), and thus provides a suboptimal approximation for saliency maps that are not histogram equalized. For this reason, and for the otherwise similar computation to AUC-Judd, we report AUC scores using the AUC-Judd implementation in the main paper. \\

\noindent\textbf{Sampling negatives in AUC-Borji and sAUC:}

The AUC-Borji score is calculated by repeatedly sampling a new set of negatives in 100 separate iterations and averaging these intermediate AUC computations together. On each iteration, as many negatives are chosen as fixations on the current image.

In the shuffled AUC (sAUC) variant, negatives are sampled at random from 10 other randomly-sampled images in the dataset (as many negatives are sampled as fixations on the current image), and the final score is also obtained by averaging over 100 trials. 

A note about naming: Riche et al. \cite{RicheICCV2013} refer to shuffled AUC as AUC-Borji, but here we make a distinction between Borji's implementation of AUC with randomly-sampled negatives, and sAUC with negatives sampled from other images \cite{BorjiICCV2013}.\\ \\

\noindent\textbf{Other AUC implementations:}

Our AUC implementations are location-based (as in \cite{NIPS2005_81,Gao2007NIPS,Harel07graph-basedvisual,Judd_2009,Tatler2005,Torralba:2006}), but other distribution-based implementations of AUC have also been used in saliency evaluation, where both ground truth and saliency inputs are continuous maps \cite{LeMeur2013}. The thresholding for computing the ROC curve can be performed on the ground truth map, the saliency map, or both \cite{engelke2013comparative}. In the first two cases, one of the maps is thresholded at different values, while the other map is thresholded at a single, fixed value (e.g. to keep 20\% of the pixels \cite{Torralba:2006,LeMeur2013}).

AUC is non-symmetric, and depending on which map is taken as the reference, different scores will be produced. A symmetric variant can be obtained by averaging two non-symmetric AUC calculations by swapping the two maps being compared \cite{engelke2013comparative}.

A recent AUC variant~\cite{wloka2016spatially} attempts to quantify spatial bias more directly instead of attempting to remove it with metrics like shuffled AUC.

A number of additional AUC implementations have been discussed and compared in \cite{BorjiICCV2013,RicheICCV2013,tseng2009quantifying}. \\

\noindent\textbf{Spearman's CC:}

 A nonlinear correlation coefficient (Spearman's CC), has also be used for saliency evaluation \cite{LeMeur2013,RicheICCV2013,toet2011computational}. Unlike Pearson's CC which takes into account the absolute values of the two distribution maps, Spearman's CC only compares the ranks of the values, making it robust under monotonic transformations. \\
 
 \noindent\textbf{Relationship between CC and NSS:}
 
 Recall that NSS is calculated as:
 
  {\scriptsize \[ NSS(P,Q^B) = \frac{1}{N}\sum_{i} \overline{P_i}\times Q^B_i \]} 
where $\overline{P}$ is the normalized saliency map, $Q^B$ is a binary map of fixations, and $N$ is the number of fixated pixels.
If the fixations are sampled instead from a fixation distribution $Q^D$, then the probability that a particular fixation at pixel $i$ is chosen is just the density $Q^D_i$. By sampling from $Q^D$, we can construct the binary map $Q^B$ (since $E(Q^B_i) = P(Q^B_i) = Q^D_i$). Over $M$ sets of samples from $Q^D$:

{\scriptsize \[ \mathbf{E}[NSS(P,Q^B)] = \frac{1}{M}\sum_{i} \overline{P_i}\times Q^D_i \]} 
Note that CC can be written as:
{\scriptsize \[CC(P,Q^D) =  \frac{1}{T}\sum_{i}\overline{P_i}\times \overline{Q^D_i}\]}
Where $T$ is the total number of pixels in the image, and both $P$ and $Q^D$ are normalized. Recall that NSS and CC both normalize by variance. Thus, NSS can be viewed as a kind of discrete approximation to CC. \\ 

\noindent\textbf{Symmetric KL:}

The standard implementation of KL that we use is non-symmetric by construction. A symmetric extension of KL as in \cite{borji2013state,li2015data} is computed as: $KL(P,Q)+KL(Q,P)$ (also see Jeffrey divergence \cite{puzicha1997non}). We use the asymmetric variant which allows the resulting KL score to be more easily interpreted, since it measures how good a saliency map prediction is at approximating the ground truth distribution. The symmetric variant is more appropriate for comparing saliency maps to each other or for computing inter-observer consistency, cases where it is not well defined what is the predicted versus ground truth distribution~\cite{wilming2011measures}. Unlike our variant, the symmetric variant penalizes false negatives and false positives equally. \\

\noindent\textbf{Other KL implementations:}

In this paper, the variant of KL we adopt would be called the \emph{image-based KL-divergence} according to \cite{kummerer2015information} since we compute the KL divergence between the saliency map and fixation map directly. This is in contrast to the \emph{fixation-based KL divergence} that is calculated by binning saliency values at fixated and nonfixated locations and computing the KL divergence of these histograms. 
Both versions of KL have been used for saliency evaluation under the metric name KL, leading to some confusion. The Supporting Information of \cite{kummerer2015information} includes a list of papers (Table S3) using each of these varieties. 
There is also a shuffled implementation of KL available \cite{Zhang2008JOV} to discount central predictions similar to shuffled AUC (sAUC). \\

 \noindent\textbf{Relationship between KL and IG:}
 
This relationship is discussed at length by K{\"u}mmerer et al. \cite{kummerer2014close,kummerer2015information}.
Here we explicate this relationship for our formulation of KL and IG. Recall that:

{\scriptsize \[ KL(P,Q^{D}) = \sum_{i}Q^{D}_{i} \log\left(\epsilon + \frac{Q^{D}_{i}}{\epsilon+P_{i}}\right) \] }
Where $i$ iterates over all the pixels in the distribution $Q^D$ (approximating an integral).
Then: 
{\scriptsize \[ KL(B,Q^{D}) - KL(P,Q^{D}) \] }
{\scriptsize \[ = \sum_{i}Q^{D}_{i} \left[\log\left(\epsilon + \frac{Q^{D}_{i}}{\epsilon+B_{i}}\right) - \log\left(\epsilon + \frac{Q^{D}_{i}}{\epsilon+P_{i}}\right)\right] \] }
which for very small $\epsilon$ approaches:
{\scriptsize \[ \sum_{i}Q^{D}_{i} \left[\log\left(\frac{\epsilon+P_{i}}{\epsilon+B_{i}}\right)\right] \] }
yielding the discrete approximation:
{\scriptsize \[ \frac{1}{N}\sum_{i}Q^{B}_{i} \left[\log\left(\frac{\epsilon+P_{i}}{\epsilon+B_{i}}\right)\right] \] }

\noindent and within a constant factor (due to change of base from natural log to base 2), this is equal to $IG(P,Q^{B})$. Information gain is measured in terms of bits/fixation.
Information gain is like KL but baseline-adjusted (recall also that KL is a dissimilarity metric, while IG is a similarity metric, explaining the change of places between $P$ and $B$). The additional distinction is that IG is more computationally similar to fixation-based KL, rather than image-based KL (which we use in the main paper). \\

\noindent\textbf{IG:}

For the IG visualizations in the paper, we compute a per-pixel value of: $V_i = \log_2(\epsilon + P_{i}) - \log_2(\epsilon + B_{i})$. This value is then modulated by the human fixation distribution $Q^{D}$. In red are all pixels where $Q^{D}_iV_i < 0$, and in blue are all pixels where $Q^{D}_iV_i > 0$. Note that the visualizations in this paper are different from the ones in \cite{kummerer2014close,kummerer2015information}.\\

\noindent\textbf{EMD:}

We use a fast implementation of EMD provided by Pele and Werman \cite{Pele-eccv2008} \cite{Pele-iccv2009}\footnote{Code at \url{http://www.cs.huji.ac.il/~ofirpele/FastEMD/}} but without a threshold. For additional efficiency, we resize both maps to 1/32 of their size after they are first resized to the same dimensions. The maps are then normalized to sum to one. Despite these modifications, EMD is more computationally expensive to compute than any of the other metrics because it requires joint optimization across all the image pixels. 

For visualization, at pixel $i$ we plot $D_{from} = \sum_{j}f_{i j}d_{i j}$ in green for all $i$ where $D_{from} > 0$, and at pixel $j$, we plot $D_{to} = \sum_{i}f_{i j}d_{i j}$ in red for all $j$ where $D_{to} > 0$. Note that the set of pixels where $D_{from}>0$ is disjoint from the set of pixels where $D_{to}>0$, so each pixel is either red or green or neither.

\subsection{Normalization of saliency maps}

Metric computations often involve normalizing the input maps. This allows maps with different saliency value ranges to be compared.
A saliency map $S$ can be normalized in a number of ways:

\vspace{1em}
\noindent\textbf{(a) Normalization by range: } 
$S \rightarrow \frac{S-min(S)}{max(S)-min(S)}$ \\ \\
\noindent\textbf{(b) Normalization by variance\footnote{This is also often called standardization.}: } 
$S \rightarrow \frac{S-\mu(S)}{\sigma(S)}$\\ \\
\noindent\textbf{(c) Normalization by sum: } 
$S \rightarrow \frac{S}{sum(S)}$ 

\vspace{1em}
\noindent Table~\ref{tab:normalization} lists the normalization strategies applied to saliency maps by the metrics in this paper. Another approach is \textbf{normalization by histogram matching}, with histogram equalization being a special case. Histogram matching is a monotonic transformation that remaps (re-bins) a saliency map's values to a new set of values such that the number of saliency values per bin matches a target distribution. 
\\

\noindent\textbf{Effect of normalization on metric behaviors:}

Histogram matching does not affect AUC calculations\footnote{Unless the thresholds for the ROC curve are not adjusted accordingly. For instance, in the ROC-Borji implementation with uniform threshold sampling, histogram matching changes the number of saliency map values in each bin (at each threshold).}, but does affect all the other metrics. Histogram matching can make a saliency map more peaked or more uniform. This has different effects on metrics: for instance, EMD prefers sparser maps provided the predicted locations are near the target locations (the less density to move, the better). However, more distributed predictions are more likely to have non-zero values at the target locations and better scores on the other metrics. These are important considerations for preprocessing saliency maps. 

\newcommand{\lw}{0.18\linewidth}
\begin{table}
\centering
\begin{tabular}{| >{\centering\arraybackslash}p{\lw} | >{\centering\arraybackslash}p{\lw} |  >{\centering\arraybackslash}p{\lw} |  >{\centering\arraybackslash}p{\lw} |}
\hline
Metric & Normalized by range & Normalized by variance & Normalized by sum \\ \hline
AUC & \checkmark & & \\ \hline
sAUC & \checkmark & & \\ \hline
NSS & & \checkmark & \\ \hline
CC & & \checkmark &  \\ \hline
EMD & & & \checkmark \\ \hline
SIM & & & \checkmark \\ \hline
KL & & & \checkmark \\ \hline
IG & & & \checkmark \\ \hline
\end{tabular}
\caption{{\small Different metrics use different normalization strategies for pre-processing saliency maps prior to scoring them. Normalization can change the extent to which the range of saliency values and outliers affect performance.}}
\label{tab:normalization}
\vspace{-1em}
\end{table}

Different normalization schemes can also change how metric scores are impacted by very high and very low values in a saliency map. For instance, in the case of NSS, if a large outlier saliency value occurs at least at one of the fixation locations, then the resulting NSS score will be correspondingly high (since it is an outlier, it will not be significantly affected by normalization). Alternatively, if most saliency map values are large and positive except at fixation locations, then the normalized saliency map values at the fixation locations can be arbitrarily large negative numbers. \\

\noindent\textbf{Normalization for conversion to a density:}

It is a common approach during saliency evaluation to take a saliency map as input and normalize it by its sum to convert it into a probability distribution, prior to computation of the SIM, KL, and IG scores. However, if the initial map was not designed to be probabilistic, this transformation is not sufficient to qualify the map as a probabilistic map. 
For instance, a value of zero in a probabilistic map implies the map predicts that fixations are impossible in this image region. This is why regularization is an important factor for probabilistic maps. Adding a small epsilon to a map's predictions can drastically improve its KL or IG score. Note additionally that if a saliency map is stored in a compressed format, small regularization values might not be preserved, so the format of the saliency map can either facilitate or hinder evaluation according to different assumptions (see Sec.~\ref{sec:recommendations} below).



\subsection{Baselines and bounds}

\noindent\textbf{Empirical limits of metrics}

One of the differences between location-based and distribution-based metrics is that the empirical limits of location-based metrics (AUC, sAUC, NSS, IG) on a given dataset do not reach the theoretical limits (Table~\ref{tab:difObsNum}).
The sets of fixated and non-fixated locations are not disjoint, and thus no classifier can reach its theoretical limit \cite{wilming2011measures}. In this regard, the distribution metrics are more robust. Although different sets of observers fixate similar but not identical locations, continuous fixation maps converge to the same underlying distribution as the number of observers increases. 
To make scores comparable across metrics and datasets, empirical metric limits can be computed. 
Empirical limits are specific to a dataset, dependent on the consistency between humans, and can be used as a realistic upper bound for model performance.

\begin{figure}
\begin{center}
\includegraphics[width=0.8\linewidth]{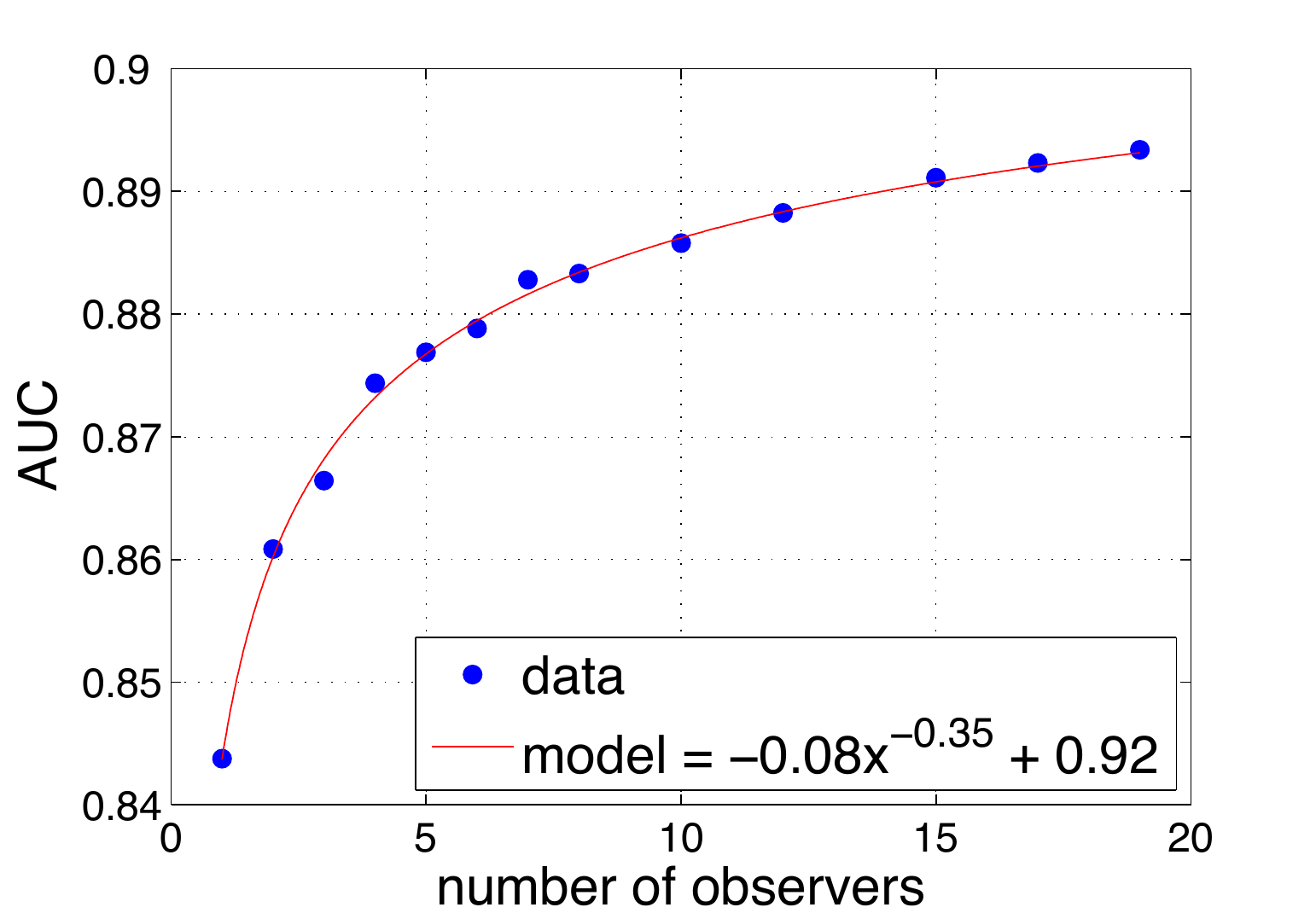}

\caption{\small{We plot the AUC-Judd scores when the fixations of $n$ observers are used to predict the fixations of another $n$ observers, for increasing $n$. Based on extrapolation of the power curve that fits the data, the limit of human performance is 0.92 under AUC-Judd. 
}} 
\vspace{-1.5em}
\label{fig:humanPerformance}
\end{center}
\end{figure}

We measured human consistency using the fixations of one group of $n$ observers to predict the fixations of another group of $n$ observers. By increasing the number of observers, we extrapolated performance to infinite observers. After computing performance for
$n=1$ to $n=19$ (half of the total 39 observers), we fit these points to the power function $f(n) = a*n^b+c$, constraining $b$ to be negative and $c$ to lie within the valid theoretical range of the metric (see Fig.~\ref{fig:humanPerformance}). The results of the fitting function\footnote{Matlab's \emph{fit} function, using non-linear least squares fitting.} that we include in Table \ref{tab:difObsNum} are the empirical limit $c$ and the 95\% confidence bounds.
Once the empirical limit has been computed for a given metric on a given dataset, this limit can be used to normalize the scores for all computational models \cite{Peters2005}.

\newcommand{\pt}{1.4cm}
\begin{table*}
\centering
\begin{tabular}{| p{2.4cm} | p{\pt} | p{\pt} | p{\pt} | p{\pt} | p{\pt} | p{\pt} | p{1cm} | p{1cm} |}
\hline
Metric limits &\multicolumn{6}{c |}{Similarity metrics}&\multicolumn{2}{c |}{Dissimilarity metrics}\\
\hline
 & AUC $\uparrow$ & sAUC $\uparrow$ & NSS $\uparrow$ & SIM $\uparrow$ & CC $\uparrow$ & IG $\uparrow$ & EMD $\downarrow$ & KL $\downarrow$ \\ \hline
Theoretical range (best score in bold) & [0,\textbf{1}] & [0,\textbf{1}] & [$-\infty$,\boldmath$\infty$] &  [0,\textbf{1}] & [-1,\textbf{1}] & [$-\infty$,\boldmath$\infty$] & [\textbf{0},$\infty$]  & [\textbf{0},$\infty$] \\ \hline
Empirical limit & 0.92 & 0.81 & 3.29 & 1& 1 &  2.50 & 0 & 0 \\
(with 95\% confidence bounds) & (0.91; 0.93) & (0.79; 0.83) & (3.08; 3.50) & (0.76; 1.24) & (0.82; 1.18) & (2.14; 2.86) & 0 & 0  \\ \hline
\end{tabular}
\caption{{\small Different metric scores span different ranges, while the empirical limits of the metrics are specific to a dataset. Taking into account the theoretical and empirical limits makes model comparison possible across metrics and across datasets. An empirical limit is the performance achievable on this dataset by comparing humans to humans. It is calculated by computing the score when $n$ observers predict another $n$ observers, with $n$ taken to the limit by extrapolating empirical data. Included are upper limits for the similarity metrics and lower limits for the dissimilarity metrics. 
}}
\label{tab:difObsNum}
\end{table*}



Other researchers compute the limit of human consistency as the inter-observer (IO) model where all other observers are used to predict the fixations of the remaining observer \cite{Borji_2012,LeMeur2013,Peters2005,wilming2011measures}. The resulting scores are usually averaged over all or a subset of observers. To avoid confusion, note that this IO model is different from our single observer model: in the IO model $n-1$ observers predict 1 observer; the single observer model, 1 observer predicts $n-1$ observers. \\ 




\noindent\textbf{Alternative baselines:}

For our \textbf{center prior} model we use a Gaussian stretched to the aspect ratio of the image. This version of the center prior performs slightly better than an isotropic Gaussian because objects of interest tend to be spread along the longer axis. See Clarke and Tatler~\cite{Clarke2014} for an analysis of different types of center models.

Our \textbf{chance} model assigns a uniform value to each pixel in the image. According to this model, there are few zero values in the resulting chance maps.
An alternative interpretation of chance could be to create a random fixation map by randomly selecting a number of locations in the image to serve as fixation locations, and Gaussian blurring the result. This chance model is likely to perform differently according to our metrics because of its different properties. In particular, the greater sparsity in the map, if not regularized properly, would lead to low KL, IG, and SIM scores.

For our \textbf{single observer} model we use the fixation map from one observer to predict the fixations of the remaining observers. We compute the single observer fixation map by Gaussian blurring the fixations of an observer with blur sigma equal to 1 degree of viewing angle in the ground truth eye tracking data. A different blur sigma or regularization factor for this model may compensate for the sparse predictions this model makes and improve its performance according to the KL, IG, and SIM metrics.
\\



\subsection{Recommendations for designing a saliency benchmark} \label{sec:recommendations}


\noindent\textbf{Regarding histogram matching:}

Prior to September 2014, the MIT Saliency Benchmark histogram matched saliency maps to a target distribution before evaluation \cite{Judd_2012}. This was intended to reduce differences in saliency map ranges. However, this had significant effects on model performances, inflating or deflating scores, depending on the model. The decision was made to evaluate saliency maps as-is and to leave any preprocessing to the model submitters. This also makes reporting more transparent, as the scores posted on the website directly correspond to the maps submitted.\\ 

\noindent\textbf{Saliency map input:}

K{\"u}mmerer et al. argues that a probabilistic definition is most intuitive for saliency models because it makes the saliency value in an image region easily interpretable: as the probability that a fixation is expected to occur there; or if differently normalized, as the expected number of fixations to occur in that region from an observer or population of observers~\cite{kummerer2014close,kummerer2015information}. It also makes the relative values in a saliency map meaningful; e.g., a region with twice the saliency value can be expected to have twice the fixations.

The precise format in which a saliency model is submitted for evaluation (i.e., to a saliency benchmark) also affects the resulting performance numbers. For instance, jpg-encoded images only save 8 bits per pixel, and the jpg artifacts can have a large impact in image regions with low saliency values. A better approach is requiring model entries in non-compressed formats. A map saved as a log probability map instead of just as a probability map is also better for representing a larger range of values, and for preserving small saliency values (e.g., for regularization). A given saliency benchmark should specify what kind of input is required, so that both the model submitters and evaluators operate under the same set of assumptions, and for the saliency map values to be handled correctly during evaluation.